\documentclass[11pt]{article}
\usepackage[margin=1in]{geometry}

\usepackage[in]{fullpage}
\usepackage{multirow}
\usepackage{graphicx}
\usepackage{amsmath}
\usepackage{bm}
\usepackage{amsfonts}
\usepackage{tikz}
\usepackage{natbib}

\usepackage{algorithm,caption}
\usepackage[noend]{algorithmic}

\usepackage[utf8]{inputenc} 
\usepackage[T1]{fontenc}    
\usepackage{url}            
\usepackage{booktabs}       
\usepackage{amsfonts,amsmath}       
\usepackage{nicefrac}       
\usepackage{microtype}      
\usepackage{xcolor}         
\usepackage{amsmath,mleftright}

\usepackage{amsfonts,amsmath,amssymb,amsthm}       
\usepackage{bm}
\usepackage{natbib}
\usepackage{graphicx}
\usepackage{subfigure}
\usepackage{hyperref}
\usepackage{cleveref}
\usepackage{bbm}
\usepackage{xspace}
\usepackage{bigints}
\usepackage{amssymb,amsopn,float,bbm,bm,enumerate,color,multirow,gensymb}

\usepackage{epsfig,subfigure,graphicx}
\usepackage{comment}
\usepackage{afterpage}
\usepackage{thmtools,thm-restate}

\usepackage[]{color-edits}

\addauthor{as}{blue}
\addauthor{rd}{purple}

\newcommand{\beq}{\begin{equation}}
\newcommand{\eeq}{\end{equation}}

\newcommand\E{\mathbb{E}}

\newcommand\1{\mathbbm{1}}

\newcommand{\cH}{{\cal H}}
\newcommand{\cI}{{\cal I}}

\DeclareMathOperator{\argmax}{argmax}

\newtheorem{defn}{Definition}[section]

\theoremstyle{definition} 

\newtheorem{remark}{Remark}[section]

\newcommand {\commentout}[1] {}

\def\ints{{{\rm Z} \kern -.35em {\rm Z} }}  
\def\smallints{{{\rm Z} \kern -.3em {\rm Z} }}  
\def\pints{{{\rm I} \kern -.15em {\rm N} }}      
\newcommand{\reals}{\mathbb R}

\def\cplx{{{\rm I} \kern -.45em {\rm C} }}       
\def\l2{\rm {\mathcal L}^{2}(\reals)}            

\newtheorem{definition}{Definition}[section]
\newtheorem{nad}{Notation and Definitions}[section]

\newcommand{\be}{\begin{eqnarray}}
\newcommand{\ee}{\end{eqnarray}}
\newcommand{\bea}{\begin{eqnarray}}
\newcommand{\eea}{\end{eqnarray}}
\newcommand{\beaa}{\begin{eqnarray*}}
\newcommand{\eeaa}{\end{eqnarray*}}
\newcommand{\bnad}{\begin{nad}}
\newcommand{\enad}{\end{nad}}

\newcommand{\cO}{\mathcal{O}}

\newcommand{\IGNORE}[1]{}

\newcommand{\optb}{\text{OPT}^{(b)}}
\newcommand{\rewb}{\text{REW}^{(b)}}

\newcommand{\regb}{\text{REG}^{(b)}}

\newcommand{\optc}{\text{OPT}^{(c)}}
\newcommand{\rewc}{\text{REW}^{(c)}}
\newcommand{\regc}{\text{REG}^{(c)}}

\newcommand{\optbx}{\text{OPT}_{x}^{(b)}}
\newcommand{\optby}{\text{OPT}_{y}^{(b)}}

﻿

\renewcommand{\E} {\operatornamewithlimits{\ensuremath{\mathbb{E}}}} 

{\begin{array}{l@{\hspace{8mm}}l@{\hspace{8mm}}l}}%
{\end{array}}
\newlength{\lplb}
\newcommand{\vig}[0]{$\mathtt{Constrained\;D}$-$\mathtt{EXP3}$}
\title{ \textbf{Think Before You Duel: Understanding Complexities of Preference Learning under Constrained Resources}}

\author{%
  Rohan Deb \\
  University of Illinois Urbana-Champaign\\
  \texttt{rd22@illinois.edu}
  \and
  Aadirupa Saha\\
  TTIC\\
  \texttt{aadirupa@ttic.edu}
}
\date{}
\begin{document}

\maketitle

\begin{abstract}
 We consider the problem of reward maximization in the dueling bandit setup along with constraints on resource consumption. As in the classic dueling bandits, at each round the learner has to choose a pair of items from a set of $K$ items and observe a relative feedback for the current pair. Additionally, for both items, the learner also observes a vector of resource consumptions. The objective of the learner is to maximize the cumulative reward, while ensuring that the total consumption of any resource is within the allocated budget. We show that due to the relative nature of the feedback, the problem is more difficult than its bandit counterpart and that without further assumptions the problem is not learnable from a regret minimization perspective. 
Thereafter, by exploiting assumptions on the available budget, we provide an EXP3 based dueling algorithm that also considers the associated consumptions and show that it achieves an $\tilde{\cO}\left({\frac{\optb}{B}}K^{1/3}T^{2/3}\right)$ regret, where $\optb$ is the optimal value and $B$ is the available budget. Finally, we provide numerical simulations to demonstrate the efficacy of our proposed method.

\end{abstract}

\section{Introduction}
\label{sec:intro}
The standard Multi-Armed Bandit (MAB) setting involves an agent learning from stochastic feedback, provided in the form of numerical rewards \citep{lai-allocation,auer,lattimore_szepesvári_2020}. At every round $t\in [T]$ the learner pulls an arm from $K$ arms and the environment provides a reward $r_t$ drawn i.i.d. from a distribution unknown to the learner. The objective of the learner is to maximize the cumulative reward over the time horizon $T$. In many real world scenarios, e.g., movie recommendations, ad placements, retail management, tournament ranking, search engine optimization, one does not receive a numerical reward but rather receives feedback in terms of pairwise comparisons or rankings. 
In the simplest \emph{Dueling Bandits} setup, at each round $t\in [T]$ the learner picks two items $i,j$ from $K$ arms, and receives the output of a duel between the two, i.e., whether item $i$ is preferred over $j$ or vice versa. The objective of the learner is to minimize regret as compared against a \emph{best} arm in hindsight. This setting has garnered a fair amount of attention over several years \citep{YUE20121538,btm,rucb,gajane15,ailon2014reducing,Zoghi+14RUCB}.
The simplicity and ease of data collection techniques of preference based learning gained huge interest in the online learning community. As a result, Dueling Bandits has been generalized and studied in multiple practical scenarios, e.g., extending pairwise preferences to subsetwise preferences \cite{SG18,Ren+18,SGrum20}, finite to large or potentially infinite decision spaces \cite{ContDB,yue2009interactively,SKM21}, stochastic to adversarial preferences \cite{gajane15,pmlr-v139-saha21a} and contextual scenarios
\cite{CDB,S21,bengs2022stochastic}, item unavailability \cite{gaillard2023one}, non-stationary preferences \cite{gupta2021optimal} or even in interdisciplinary fields of research like robotics \cite{bhatia2020preference} and assortment optimization \cite{assort-mnl,assort-mallows}.

However, in many of these real-world scenarios, the agent has to minimize regret while operating under certain constraints, e.g., a limited supply of resources or a cost associated with each item. For e.g., ad placement is often constrained by the available advertiser budget and user reach, movie recommendations might have a cost associated with each recommendation, and retail management might need to worry about logistical or supply constraints. In recent times, preference based feedback is also used to train more complex systems like assistive robots and autonomous cars which are  computationally demanding and often resource constraints may limit the available resources. Therefore from the perspective of actually deploying Dueling bandit algorithms in real-world, it is essential to study the problem in a more general constrained setup - \emph{Constrained Dueling Bandits}.



\textbf{Informal Problem Setup: } At every round, the learner picks arms $x_t, y_t \in [K]$ to duel, where $K$ is the number of arms, and observes a Bernoulli output with parameter $P(x_t,y_t)$. Here $P(x_t,y_t)$ measures the probability of $x_t$ being preferred over $y_t$ and therefore the matrix $P = [P(i,j)]_{i,j \in [K]}$ is called the preference matrix. Further it also observes some consumption of resources associated with arms $x_t$ and $y_t$. The objective of the learner is to minimize the regret over a time horizon while ensuring that the total sum of consumptions is less than a pre-defined budget (see Section~\ref{sec:prelim} for a formal description). 

\subsection{Our Contributions}
We provide an outline of our main contributions here. Note in the dueling setup, the quality of an item is relative, so to estimate the quality of one arm we need to compare it with the rest. The primary challenge lies in ensuring we can actually query all the pairs, while staying within the budget constraint. A straightforward extension of standard algorithms from Bandit with Knapsack fails in DB setting because they draw arms in an UCB manner which leads to selecting the same arm twice, thus revealing no statistical information. Further existing elimination based algorithms for DB \citep{YUE20121538,btm,rucb} cannot be extended to the constrained setting since once an arm is eliminated, an unbiased estimate of the corresponding scores cannot be obtained (also see Remark~\ref{rem:5.1}). Precisely our contributions are as follows.
\begin{enumerate}
    \item \textbf{Formulation:} We setup the \emph{Constrained Dueling Bandits}(Constrained-DB) problem by defining two kinds of benchmarks, corresponding to two types of best arms/winners - \emph{Condorcet winner} and the \emph{Borda winner}, such that the benchmarks also satisfy the given constraints (see Section~\ref{sec:prelim}).
    
    \item \textbf{Lower Bounds:} We show that the `relative' nature of feedback makes Constrained-DB a difficult problem to solve (in comparison to its MAB counterpart). Specifically, we provide lower bound results for both Condorcet Constrained-DB and Borda Constrained-DB and show that the most general setup has a regret of $\Omega(T)$ and therefore one needs to impose additional assumptions either on the structure of the preference matrix or the available budget to give meaningful regret bounds (see Section~\ref{sec:Lower Bound}).
    
    \item \textbf{Algorithms and Upper Bounds:} Under assumptions on the available budget, we 
    provide an EXP3 based algorithm \vig that also take into account the associated resource consumptions before choosing two arms to duel. Thereafter we show it achieves sub-linear regret (see Section~\ref{sec:algo}).

    \item \textbf{Empirical Evaluations:} We also evaluate our proposed algorithms empirically on synthetic and real data and show that they outperform the existing DB algorithms when there are budget constraints associated with arm selections (see Section~\ref{sec:exp}).

\end{enumerate}

\section{Related Works}
\label{sec:related_works}

We briefly discuss some related literature here; for a more detailed discussion see Appendix~\ref{sec:App_related}.


\textbf{Dueling Bandits:}
The Dueling Bandits setting has seen a lot of development in the past decade. The problem in its current form was introduced in \citet{YUE20121538} and upper and lower bounds on the regret were provided by assuming that the preference matrix had some specific structures such as total ordering, strong stochastic transitivity and strong triangle inequality (also see Section~\ref{sec:prelim} for definitions). Subsequently \citep{btm} proposed `Beat the Mean' algorithm with improved regret bound while also relaxed the strong stochastic transitivity assumption to relaxed stochastic transitivity. \citet{rucb} further relaxed the total ordering assumption to the existence of a Condorcet winner (an arm that beats every other arm) and provided a upper confidence bound (UCB) based algorithm. \citep{ailon} studied the dueling bandit problem in an adversarial setup (where the preference matrix $P$ changes over time), introducing the sparring EXP3 idea, albeit without regret guarantees. Subsequent works~\cite{gajane15, pmlr-v139-saha21a} utilized this concept to prove regret guarantees in adversarial environments. 
\vspace{6pt}

\textbf{Constrained Bandits:}
There is a body of literature that under the name \emph{Bandits with Knapsacks} that looks at cumulative reward maximization under budget constraints. It was first introduced in \citep{Badanidiyuru_2013} for the MAB setting and the proposed algorithms- BalancedExploration and PrimalDualBwK were shown to enjoy optimal regret bounds up to polylogarithmic factors. BalancedExploration however is not an efficient algorithm (see Remark 4.2 in \citep{Badanidiyuru_2013}) and we do not pursue this. We further show that in fact PrimalDualBwK attains an $\Omega(T)$ regret in the MAB setting, and as such we do not try to adapt this algorithm to the Constrained-DB setting. \citep{linCBwK} study the natural extension of the problem - the linear contextual bandits with knapsacks and provide an algorithm utilizing ideas from UCB and primal-dual methods with sub-linear regret bound. Further extensions have been studied such as regret minimization with concave reward and convex objective \citep{pmlr-v49-agrawal16,agarwal14}, adversarial setting \citep{Immorlica} and smoothed adversarial setting \citep{pmlr-v162-sivakumar22a} that all use a version of the central primal-dual idea from \citep{linCBwK}. In Section~\ref{sec:algo} we discuss why a direct extension of the algorithm from \citep{linCBwK} does not work in the Constrained-DB setting (see Remark~\ref{rem:5.1}).

\section{Preliminaries}
\label{sec:prelim}

\subsection{Existing Concepts from Dueling Bandits}

A learner at round $t \in [T]$ is presented with $K$ arms to choose from. It then selects two arms $x_t, y_t \in [K] := \{1,2,\ldots,K\}$ to duel, and receives a feedback $o_t \sim \text{Ber}(P(x_t,y_t))$. Here, $\text{Ber}(p)$ denotes the Bernoulli distribution with parameter $p$ and $P(x_t,y_t)$ measures the probability of $x_t$ being preferred over $y_t$, that satisfies $P(x,y) = 1 - P(y,x)$ for all $x,y \in [K]$. Further we define the matrix $P = [P(i,j)]_{i,j \in [K]}$ and call it the preference matrix.

Next we consider two notions of winners in the dueling setup, namely the \emph{Borda winner} and the \emph{Condorcet} winner. We define them as follows.

\begin{definition}[\textbf{Condorcet Winner} \cite{rucb}]
    We define the Condorcet winner $x^{(c)}$ as the arm that is preferred over all the other arms, i.e., $x^{(c)} = i$ iff $P(i,j) > 1/2,\; \forall j \in [K]\setminus\{i\}$. Further the Condorcet score of arm $x$ is defined as
    \begin{align*}
        c(x) = P(x,x^{(c)}).
    \end{align*}
\end{definition}

\begin{definition}[\textbf{Borda Winner}\cite{pmlr-v139-saha21a}]
    We define the \emph{Borda score} of an arm $x \in [K]$ as
\begin{align*}
    b(x) = \frac{1}{K-1} \sum_{x \neq y} P(x,y)
\end{align*}
The \emph{Borda winner} $x^{(b)}$ is defined as the arm that maximizes the Borda score, i.e., $x^{(b)} = \argmax_{x} b(x)$.
\end{definition}

\begin{definition}[\textbf{Total Ordering} \citep{YUE20121538,btm}]
\label{def:totalordering}
    We say the preference matrix $P$ satisfies Total Ordering(TO) if there exists a binary total order relation $\succ$ with $\forall\; i,j \in [K]$, $i \succ j$ implies $P(i,j) > \frac{1}{2}$. 
\end{definition}
\begin{definition}[\textbf{Strong Stochastic Transitivity} \citep{YUE20121538,btm}]
\label{def:sst}
    We say the preference matrix $P$ satisfying TO condition further satisfies Strong Stochastic Transitivity (SST) if for every $i,j,k \in [K]$, $i\succ j \succ k$ implies $P(i,k) \geq \max\{P(i,j),P(j,k)\}$ where $\succ$ is the underlying TO relation.
\end{definition}



\subsection{Our Problem Setup}

While the problem of DB is well studied over the past two decades (see \cref{sec:related_works}), the literature lacks the practical setup of considering the aspects resource constraints in Dueling Bandits, which is often realizable in practical scenarios as we motivated in section~\ref{sec:intro}. In this section, we define the constrained dueling bandit setup.



\vspace{3pt}
\textbf{Constrained Dueling Bandits.} At round $t \in [T]$ when the learner selects two arms $x_t, y_t \in [K]$ to duel, it also observes two consumption vectors $u(x_t), v(y_t) \in [0,1]^d$ associated with the pulled arms $x_t$ and $y_t$, drawn independent of the past {history} from an unknown distribution. The $d$ elements of the vector are the consumptions associated with $d$ different types of resources. We define $u^*(\cdot) $ and $v^*(\cdot)$ as the expected consumptions of the two arms respectively, i.e., $u^*(x) = \E[u(x)], $ and $ v^*(x) = \E[v(x)]$. 
The learner also has the option of not choosing any arm at a round and see no feedback and incur no consumption.
The total budget available to the learner is $B \leq T$ and the interaction with the environment ends at $t=\tau$ when either $\tau = T$ (end of time horizon) or $ \left(\sum_{t=1}^{\tau} u_t(x_t) + v_t(x_t)\right)_i > B$ for some $i \in [d]$ (the budget of some resource is exhausted).

\vspace{3pt}

\textbf{Benchmarks.} Next we describe the benchmarks in the two settings that our algorithms compete against. Suppose $\pi_x(\cdot)$ and $\pi_y(\cdot)$ represent the distribution of arms played in the first selection and second selection of the dual respectively. Then the two optimal solutions are defined as.
\begin{enumerate}
    \item \textbf{Condorcet Optimal Solution} Consider the following Linear program (LP) with Condorcet score.
    \begin{equation}
        \begin{split}
            \max_{\pi_x,\pi_y \in \Delta^{K}} &\sum_{x,y \in [K]}\, \pi_x(x)\, c(x) + \pi_y(y) c(y)~,\\
        \text{such that} &\sum_{x,y\in [K]} \pi_x(x)u^*(x) + \pi_y(y)v^*(y) \leq \frac{B}{T} \mathbbm{1}~.
        \end{split}
        \label{lp:primal-Condorcet}
        \tag{$\mathtt{LP-Condorcet}$}
    \end{equation}
    where $\Delta^{K}$ is the probability simplex over $K$. Suppose $\pi_x^{*(c)},\pi_y^{*(c)}$ solve \eqref{lp:primal-Condorcet} then we define the optimal value as 
    $$\displaystyle \optc = T\sum_{x,y \in [K]}\, \pi_x^{*(c)}(x)\, c(x) + \pi_y^{*(c)}(y)\, c(y).$$

    \item \textbf{Borda Optimal Solution} Consider the following Linear program (LP) with Borda score.
    \begin{equation}
        \begin{split}
        \max_{\pi_x,\pi_y \in \Delta^{K}} &\sum_{x,y \in [K]}\, \pi_x(x)\, b(x) + \pi_y(y) b(y)~,\\
        \text{such that} &\sum_{x,y\in [K]} \pi_x(x)u^*(x) + \pi_y(y)v^*(y) \leq \frac{B}{T} \mathbbm{1}~.
        \end{split}
        \label{lp:primal-Borda}
        \tag{$\mathtt{LP-Borda}$}
    \end{equation}
where $\Delta^{K}$ is the probability simplex over $K$. Suppose $\pi_x^{*(b)},\pi_y^{*(b)}$ solve \eqref{lp:primal-Borda} then we define the optimal value as $$\displaystyle \optb = T\sum_{x,y \in [K]}\, \pi_x^{*(b)}(x)\, b(x) + \pi_y^{*(b)}(y)\, b(y).$$
\end{enumerate}

\begin{remark}
    Note that the benchmarks \eqref{lp:primal-Condorcet} and \eqref{lp:primal-Borda} compute a non-adaptive policy and follows the development in \citep{linCBwK}. It can be shown that $\optc$ and $\optb$ upper bounds the value of the corresponding optimal adaptive policy (e.g., Lemma 1 in \cite{linCBwK}).
\end{remark}

\textbf{Regret} Next we define the total cumulative regret of an algorithm that chooses the sequence of arms $\{(x_t,y_t)_{t=1}^\tau\}$, where $\tau \leq T$ is the stopping time of the algorithm. We define the following types of regret.
\begin{enumerate}
 \item \textbf{Condorcet Regret.} We define the cumulative Condorcet reward until the stopping time $\tau$ as
    \begin{align}
        \label{eq:reward_condorcet}
        \rewc = \sum_{t=1}^{\tau} c(x_t) + c(y_t),
    \end{align}
     and the corresponding Condorcet regret as
     \begin{align}
        \label{eq:regret_condorcet}
        \regc(T) = \optc - \rewc
     \end{align}
    \item \textbf{Borda Regret.} We define the cumulative Borda reward until the stopping time $\tau$ as
    \begin{align}
        \label{eq:reward_borda}
        \rewb = \sum_{t=1}^{\tau} b(x_t) + b(y_t),
    \end{align}
     and the corresponding Borda regret as
     \begin{align}
        \label{eq:regret_borda}
        \regb(T) = \optb - \rewb
     \end{align}
\end{enumerate}

\textbf{Objective} Informally the objective is to maximize the total sum of rewards while satisfying the budget constraint. Formally, the algorithm competes with the benchmarks defined in \eqref{lp:primal-Condorcet} and \eqref{lp:primal-Borda} and the performance is measured via the regret defined in \eqref{eq:regret_condorcet} and \eqref{eq:regret_condorcet}.

\textbf{Additional Notation.} For ease of exposition, we shall hide dependencies on constants and work with order notation. Towards that, we shall use the notation $n_0 = \cO(t)$ to imply that there exists constant $c$ (independent of $t$) such that $\leq n_0 \leq c t $. 
The notation $\tilde{\cO}(t)$ has a similar meaning but hides the dependence on logarithmic terms.
Further,  $n_0 = \Omega(t)$ implies there exists $c$ such that $n_0 \geq c t$ and $n_0 = {o}(t) $ implies $ \lim \frac{n_0}{t} \rightarrow 0$ .

\section{Lower Bounds}
\label{sec:Lower Bound}
{We first analyze the lower bound for the Constrained-DB problem to analyze the problem complexity and achievable regret performance. }
Detailed proofs given in Appendix~\ref{sec:App_regret_lower}.
\subsection{Lower Bounds for Condorcet Constrained-DB}
We state two lower bound results.

(1) \emph{Lemma~\ref{lemma:condorcet_lower_bound_1} states that in the most general setting, if the allocated budget $\displaystyle B = o\left( \frac{K}{{\epsilon^{(c)}_{\min}}^2}\right)$ then the regret of any algorithm for Condorcet-constrained-DB is $\Omega(T)$, where 
${{\epsilon^{(c)}_{\min}}}$ is the minimum gap in the Condorcet scores. Further the bound does not improve even if we assume that the preference matrix satisfies total ordering but does improve if we further assume that the preference matrix satisfies strong stochastic transitivity.}

(2) \emph{Lemma~\ref{lemma:condorcet_lower_bound_2} states that if the budget $\displaystyle B = o\left( \frac{K}{{\epsilon^{(c)}_{\min}}}\right)$ then any algorithm for Condorcet-constrained-DB has regret $\Omega(T)$.}

\begin{remark}
    The $\Omega(T)$ regret in (2) cannot be improved  with any structural assumptions on the preference matrix (such as TO or SST) and a similar result can be shown to hold in the Constrained MAB setting. The $\Omega(T)$ regret in (1) is far more interesting because as we will show in the sequel, this arises precisely because of the interplay between \emph{relative feedback} and \emph{budget constraints} and could potentially be improved either by assuming some structure in the preference matrix (specifically SST in this case) or that the agent has enough budget $B \geq \frac{K}{\epsilon_{\min}^2}$.
\end{remark}
\begin{restatable}{lemm}{lemmaCWLBone}
\label{lemma:condorcet_lower_bound_1}
        Consider the Constrained-DB setting with preference matrix $P$ and define the minimum gap in Condorcet scores $\displaystyle \epsilon^{(c)}_{\min} := \min_{i,j\in[K]}(|c(i)-c(j)|)$. Suppose the available budget $ B = o\left( \frac{K}{{\epsilon^{(c)}_{\min}}^2}\right)$ then there exists a preference matrix $P$ such that $\regc(T) = \Omega(T)$. Further we show that our $\Omega(T)$ regret bound exists even when $P$ satisfies total ordering (cf. Definition~\ref{def:totalordering}) but not when $P$ satisfies strong stochastic transitivity (cf. Definition~\ref{def:sst})
\end{restatable}

\emph{Proof sketch.}
We outline the idea behind the creation of the lower bound example here. For ease of exposition we consider a simplified setup with $K=3$. We start with the general setting without any assumption on the preference matrix and subsequently consider total ordering and strong stochastic transitivity.

\textbf{General setting:} Suppose the preference matrix $P$ is given by
\begin{align*}
    P &= 
    \begin{bmatrix}
        \dfrac{1}{2} & \dfrac{1}{2} + \epsilon & \dfrac{1}{2} + 2\epsilon\\[6pt]
        \dfrac{1}{2} - \epsilon & \dfrac{1}{2} & \dfrac{1}{2}\\[6pt]
        \dfrac{1}{2} - 2\epsilon & \dfrac{1}{2} & \dfrac{1}{2}
    \end{bmatrix} 
\end{align*}
and the true consumption of the three arms are given by $ u^*(1) = v^*(1) = 1, u^*(2) = v^*(2)= 0$ and $u^*(3) = v^*(3) = 0$.

The optimal policy plays arms $x_t = 1, y_t = 1$ for $B$ rounds and thereafter plays arms $x_t = 2, y_t = 2$ for the remaining $T-B$ rounds. The total accumulated reward by the optimal policy is $\optc = B + (T-B)\Big(\frac{1}{2} - \epsilon\Big)$. 

Note that arm 1 is the Condorcet winner and any algorithm needs to play the pairs (1,2) and (1,3) at least $\frac{1}{\epsilon^2}$ number of times to determine if $c(2) > c(3)$. However since the budget $B = o\left(\frac{1}{\epsilon^2}\right)$, no algorithm can determine if $c(2) > c(3)$ or $c(2) < c(3)$ and hence would always end up playing the sub-optimal arm at least $\frac{(T-B)}{2}$ number of times after the initial $B$ rounds. Therefore 
\begin{align*}
    \rewc \leq B + \frac{(T-B)}{2}\Big(\frac{1}{2} - \epsilon\Big) + \frac{(T-B)}{2}\Big(\frac{1}{2} - 2\epsilon\Big)
\end{align*}
which implies $\optc \geq \frac{(T-B)}{2}\epsilon = \Omega(T)$

The key observation here is that although arm 2 and arm 3 have zero consumption playing the pair $(2,3)$ does not provide any information about whether $s(2) > s(3)$ or $s(3) > s(2)$. This is in contrast to the standard MAB setting where playing arms 2 and 3 does give information about whether $s(2) > s(3)$ or $s(2) < s(3)$.

\textbf{Total Ordering:} Next consider the preference matrix $P$ below that satisfies total ordering.
\begin{align*}
    P &= 
    \begin{bmatrix}
        \dfrac{1}{2} & \dfrac{1}{2} + \epsilon & \dfrac{1}{2} + 2\epsilon\\[6pt]
        \dfrac{1}{2} - \epsilon & \dfrac{1}{2} & \dfrac{1}{2}+\epsilon\\[6pt]
        \dfrac{1}{2} - 2\epsilon & \dfrac{1}{2}-\epsilon & \dfrac{1}{2}
    \end{bmatrix} 
\end{align*}
with the same consumptions as before. Does playing the pair $(2,3)$ give us any information about $s(2) > s(3)$? The answer is no. This is because we have another instance with 
\begin{align*}
    P' &= 
    \begin{bmatrix}
        \dfrac{1}{2} & \dfrac{1}{2} + \epsilon & \dfrac{1}{2} + 2\epsilon\\[6pt]
        \dfrac{1}{2} - \epsilon & \dfrac{1}{2} & \dfrac{1}{2}-\epsilon\\[6pt]
        \dfrac{1}{2} - 2\epsilon & \dfrac{1}{2}+\epsilon & \dfrac{1}{2}
    \end{bmatrix} 
\end{align*}


such that although $s(2) > s(3)$, in the total ordering sense $3\succ2$. Therefore the algorithm cannot distinguish between the instances with preference matrices $P'$ and $P'$

\textbf{Strong Stochastic Transitivity:} 
Finally suppose we assume that the Preference matrix follows strong stochastic transitivity. With this assumption, notice that the instance $P'$ is not allowed and therefore the algorithm may learn about $s(2)>s(3)$ from the total order relation $2 \succ 3$ by playing the pair $(2,3)$ without consuming any resources.
\hfill \qed

The next Lemma shows that if the available budget is less than $\frac{K}{\epsilon_{\min}}$ then the regret of any algorithm is $\Omega(T)$.
\begin{restatable}{lemm}{lemmaCWLBtwo}
\label{lemma:condorcet_lower_bound_2}
        Consider the Constrained-DB setting with preference matrix $P$ and define the minimum gap in Condorcet scores $\displaystyle \epsilon^{(c)}_{\min} := \min_{i,j\in[K]}(|c(i)-c(j)|)$. Suppose the available budget $\displaystyle B = o\left( \frac{K}{\epsilon^{(c)}_{\min}}\right)$ then there exists a preference matrix $P$ such that $\regc(T) = \Omega(T)$.
\end{restatable}


\subsection{Lower Bounds for Borda Constrained-DB}

We state similar lower bounds for the Borda Constrained Dueling Bandits setting.

\begin{restatable}{lemm}{lemmaBWLBone}
\label{lemma:borda_lower_bound_1}
        Consider the Constrained-DB setting with preference matrix $P$ and define the minimum gap in Borda scores $\displaystyle \epsilon^{(b)}_{\min} := \min_{i,j\in[K]}(|b(i)-b(j)|)$. Suppose the available budget $ B = o\left( \frac{K}{{\epsilon^{(b)}_{\min}}^2}\right)$ then there exists a preference matrix $P$ such that $\regb = \Omega(T)$. 
\end{restatable}
The above lower bound could potentially be removed by assuming total ordering.
\begin{restatable}{lemm}{lemmaBWLBtwo}
\label{lemma:bbbborda_lower_bound_2}
        Consider the Constrained-DB setting with preference matrix $P$ and define the minimum gap in Borda scores $\displaystyle \epsilon^{(c)}_{\min} := \min_{i,j\in[K]}(|b(i)-b(j)|)$. Suppose the available budget $\displaystyle B = o\left( \frac{K}{\epsilon^{(b)}_{\min}}\right)$ then there exists a preference matrix $P$ such that $\regb = \Omega(T)$.
\end{restatable}
The above lower bound cannot be removed even by assuming total ordering or strong stochastic transitivity.

\section{Algorithm and Regret Bound}
\label{sec:algo}


We consider the Borda Constrained Dueling Bandits under the assumption that the given budget $B = \cO\left(\max\{\frac{K}{\epsilon_{\min}^2}, T^{3/4}\}\right)$. Our algorithm Vigilant D-EXP3 (Dueling EXP3) is outlined in Algorithm~\ref{alg:vig-dexp3}. Before proceeding to a detailed description of our algorithm, we modify the Borda benchmark in two ways that do not change the benchmark value $\optb$ by more than a constant factor.
\begin{enumerate}
\item \textbf{Shifted Borda Score \citep{pmlr-v139-saha21a}} We replace the Borda score $b(x)$ in \eqref{lp:primal-Borda} by the shifted Borda score $\tilde{b}(x)$ defined below.
\begin{defn}[\textbf{Shifted Borda Score}]
\label{rem:regeqv}
The {shifted Borda score} of item $i \in [K]$ is given by $$\displaystyle \tilde{b}(x) := \frac{1}{K} \sum_{j \in [K]} P(i,j).$$
\end{defn}
\textbf{Benchmark with shifted Borda score.} We define the benchmark LP with shifted Borda score as 
 \begin{align}
    \max_{\pi_x,\pi_y \in \Delta^{K}} &\sum_{x\in [K]} \pi_x(x) \tilde{b}(x)  + \sum_{y\in [K]} \pi_y(y)\tilde{b}(y) \nonumber \\
    \text{such that } &\sum_{x,y \in [K]} \pi_x(x) u^*(x)  +  \pi_y(y) v^*(y)  \leq \frac{B}{T} \mathbf{1}
    \tag{$\mathtt{LP-Shifted\text{-}Borda}$}
    \label{eq:LP-Shifted-Borda}
\end{align}
Let the solution to the above LP be $\widetilde{\pi}^{*(b)}$ and let
\begin{align}
\label{eq:opt-tilde}
    \widetilde{\optb} &= T\sum_{x,y \in [K]}\, \widetilde{\pi}^{*(b)}_{x}\, \tilde{b}(x) + \widetilde{\pi}^{*(b)}_{y}\, \tilde{b}(y)\\
\label{eq:rew-tilde}
    \widetilde{\rewb} &= \sum_{t=1}^{\tau} \tilde{b}(x_t) + \tilde{b}(y_t).
\end{align}
Then in the following lemma we show that the original optimal value and the total reward is constant times the optimal value and total reward respectively, with shifted Borda score.

\begin{restatable}{lemm}{lemmaoptshifted}
\label{lemm:lemmaoptshifted}
For $\widetilde{\optb}$ and $\widetilde{\rewb}$ as defined in \eqref{eq:opt-tilde} and \eqref{eq:rew-tilde} we have $$\regb = \optb - \rewb \leq \frac{K}{K-1}(\widetilde{\optb} - \widetilde{\rewb}).$$
\end{restatable}
Therefore bounding $\widetilde{\optb}-\widetilde{\rewb}$ bounds the final regret $\regb$ upto a constant factor of $\frac{K}{K-1}$.

\item \textbf{Define Separate LPs.} Next we relax the benchmark \eqref{eq:LP-Shifted-Borda} by separating the LPs associated with the two arms as defined below.
\begin{align}
    \max_{\pi_x \in \Delta^{K}} &\sum_{x\in [K]} \pi_x(x) \tilde{b}(x)\nonumber\\
    \text{such that } &\sum_{x \in [K]} \pi_x(x) u^*(x) \leq \frac{B}{2T} \mathbf{1}
    \tag{$\mathtt{LP-Shifted\text{-}Borda\text{-}x}$}
    \label{eq:LP-Shifted-Borda-x}
\end{align}
\begin{align}
    \max_{\pi_y \in \Delta^{K}} &\sum_{y\in [K]} \pi_y(y) \tilde{b}(y)\nonumber\\
    \text{such that } &\sum_{y \in [K]} \pi_y(y) v^*(y) \leq \frac{B}{2T} \mathbf{1}
    \tag{$\mathtt{LP-Shifted\text{-}Borda\text{-}y}$}
    \label{eq:LP-Shifted-Borda-y}
\end{align}
Following lemma proves that the sum of optimal values of the two LPs upper bounds the value of the optimal policy $\widetilde{\optb}$.

\begin{restatable}{lemm}{lemmaoptmarginal}
\label{lemm:lemmaoptmarginal}
    Let the optimal value of \eqref{eq:LP-Shifted-Borda-x} and \eqref{eq:LP-Shifted-Borda-y} be $\widetilde{\optbx}$ and $\widetilde{\optby}$. Then
    \begin{align*}
        \widetilde{\optbx} + \widetilde{\optby} \geq \widetilde{\optb}.
    \end{align*}
\end{restatable}
Therefore bounding $\displaystyle \widetilde{\optbx} - \sum_{t=1}^{\tau}\tilde{b}(x_t)$ and $\displaystyle \widetilde{\optby} - \sum_{t=1}^{\tau}\tilde{b}(y_t)$ separately bounds the final regret $\widetilde{\optb} - \widetilde{\rewb}$.
\end{enumerate}

 \begin{algorithm}[t!]
 \caption{\vig}
   \label{alg:vig-dexp3}
\begin{algorithmic}[1]
   \STATE {\bfseries Input:} Item set indexed by $[K]$, learning rate $\eta > 0$, exploration parameter $\gamma \in (0,1),$ and $\cO(\frac{\optb_w}{B})\leq Z_w \leq \cO(\frac{\optb_w}{B} + 1)\;, w\in \{x,y\}$ 
   \STATE {\bfseries Initialize:} Initial probability distribution $q^x_1(i) = q^y_1(i) = 1/K, ~\forall i \in [K]$
   \FOR {$t = 1, \ldots, T$}
   \STATE Sample $x_t\sim q^x_t, y_t \sim q^y_t$ i.i.d.
   \STATE Receive preference $o_{t}(x_t,y_t) \sim \text{Ber} (P_t(x_t,y_t))$ and the consumption vectors $u_t(x_t)$ and $v_t(x_t)$.
   \STATE Estimate the shifted Borda scores and the consumption vectors, for all $i \in [K]$: 
   \begin{align*}
       \hat{b}_t(i) &= 
       \frac{\1(x_t = i)}{K q^x_t(i)}\sum_{j \in [K]}\frac{\1(y_t = j)o_t(x_t,y_t)}{q^y_t(j)},\\
        \hat{u}^x_t (i) &= 1 - \frac{\1(x_t = i)}{q_t(i)} (1-u_t(x_t)),\\
        \hat{u}^y_t (i) &= 1 - \frac{\1(y_t = i)}{q_t(i)} (1-v_t(y_t)).
   \end{align*}
   \STATE Estimate the Lagrangians $\forall\; i \in [K]$
   \begin{align*}
       \hat{\ell}^x_t(i) &= \hat{b}_t(i) + Z_x \lambda_{t}^{x \top} \bigg[\frac{B}{2T}\mathbbm{1} - \hat{u}^x_t(i)\bigg]\\
       \hat{\ell}^y_t(i) &= \hat{b}_t(i) + Z \lambda_{t}^{y \top} \bigg[\frac{B}{2T}\mathbbm{1} - \hat{v}_t(i)\bigg]
   \end{align*}
   \STATE Update for all $i \in [K]$: 
   \begin{align*}
        \tilde{q}^x_{t+1}(i) &= \dfrac{\exp(\eta_x \sum_{s = 1}^t \hat{\ell}^x_s(i))}{\sum_{j = 1}^{K} \exp(\eta_x \sum_{s = 1}^t \hat{\ell}^x_s(j))}\\
       q^x_{t+1}(i) &= (1-\gamma_x)\tilde{q}^x_{t+1}(i) + \frac{\gamma_x}{K}
   \end{align*}
   \begin{align*}
       \tilde{q}^y_{t+1}(i) &= \dfrac{\exp(\eta_y \sum_{s = 1}^t \hat{\ell}^y_s(i))}{\sum_{j = 1}^{K} \exp(\eta_y \sum_{s = 1}^t \hat{\ell}^y_s(j))}\\
       q^y_{t+1}(i) &= (1-\gamma_y)\tilde{q}^y_{t+1}(i) + \frac{\gamma_y}{K}
   \end{align*}
    \STATE Update $\lambda^x_t$ and $\lambda^y_t$ using any online convex optimization on the following objective functions 
    $$g^x_{t}(\lambda) = \Big \langle \frac{B}{2T} \mathbbm{1} - \hat{u}^x_t(x_{t})\Big \rangle , \; g^y_{t}(\lambda) = \Big \langle \frac{B}{2T} \mathbbm{1} - \hat{v}^x_t(y_{t})\Big \rangle.$$
   \ENDFOR
\end{algorithmic}
\end{algorithm}

In Algorithm~\ref{alg:vig-dexp3} we maintain two distributions $q^{x}_{t}$ and $q^{y}_{t}$ to sample the arms $x_t$ and $y_t$ at time $t$. Initially both distributions are initialized to the uniform distribution (Step 2). At time $t \in [T]$ the arms $x_t$ and $y_t$ are sample from the distributions $q^{x}_{t}$ and $q^{y}_{t}$ respectively and observe the preference output $o_{t}(x_t,y_t) \sim \text{Ber}(P_t(x_t,y_t))$ and the consumption vectors $u_t(x_t)$ and $v_t(x_t)$ (Step 4 and 5). Next we compute unbiased estimates of the shifted Borda score and the two consumption vectors (Step 6) and the empirical lagrangians for the two arms (Step 7). Next we update the arm distributions $q^x_t$ and $q^y_t$ using exponential weights on
the estimated cumulative lagrangians along with an $\gamma$-uniform exploration (Step 8). Finally we update the lagrange multipliers on the dual objectives (Step 9).
\begin{remark}[\textbf{Overcoming Challenges}]
\label{rem:5.1}
    Although we follow the primal dual-approach from \citep{pmlr-v49-agrawal16,agarwal14,pmlr-v162-sivakumar22a}, we do not construct a UCB estimate of the lagrangian (by constructing UCB estimates of the rewards and LCB estimates of the consumptions) and draw the the arm optimistically. This is because in the Dueling setting, such an approach would lead to choosing the same arm twice which does not reveal any statistical information since $P(i,i) = 1/2, \; \forall i \in [K]$ is already known. Further the approach fom RUCB \cite{rucb}, Beat the mean \cite{btm} or Interleaved filter \cite{YUE20121538} cannot be extended to the constrained setting since these algorithms are elimination algorithms, and once an arm is eliminated, an unbiased estimate of the Borda score cannot be constructed which is essential to do a trade-off between the Borda score and the consumptions of associated arms.
\end{remark}
\begin{remark}[\textbf{Unknown $Z$}]
        Note that although our algorithm assumes that $OPT^{(b)}_x$ and $OPT^{(b)}_y$ are known, in the case they are unknown, the standard trick of estimating $Z_x$ and $Z_y$ for the first $\mathcal{O}(\sqrt{T})$ rounds can be used (see e.g., Section 3.3 in \citep{linCBwK}).
\end{remark}

\begin{restatable}{theo}{theoremmain}
\label{theorem:main} For $\eta_x = \big(\frac{\log K}{T\sqrt{K}}\big)^{2/3}\frac{1}{2Z_x+1}, \eta_y = \big(\frac{\log K}{T\sqrt{K}}\big)^{2/3}\frac{1}{2Z_y+1}$ and $\gamma_x = \sqrt{\eta_x K}, \gamma_y = \sqrt{\eta_y K} $, w.p. $1 - \cO(\frac{1}{T^2})$ the regret of \vig\; is bounded by
    \begin{align*}
        \regb(T) \leq \tilde{\mathcal{O}}\Big(\frac{OPT}{B} (K\log K)^{1/3} T^{2/3} \Big)
    \end{align*}
\end{restatable}

\emph{Proof sketch} Here we briefly outline the steps of the proof. For details see Appendix~\ref{sec:App_regret_proof}.

\textbf{Step 1:} We use an EXP-3 kind guarantee for the first arm to get the following bound for all $a \in [K]$,
        $$\sum_{t=1}^{\tau} \hat{\ell}^x_t(a) - \sum_{t=1}^{\tau}\sum_{a} \tilde{q}^x_t(a)\hat{\ell}^x_t(a)  \leq \frac{\log K}{\eta_x} + \eta_x \sum_{t=1}^{\tau}\sum_{i=1}^{K} \tilde{q}^x_t(i)(\hat{\ell}^x_t(i))^2.$$ 
Since $\displaystyle \tilde{q}^x_t(i) = \frac{q^x_t(i) - \frac{\gamma_x}{K}}{1-\gamma_x}$,
\begin{align}
    \forall a, \; &{(1-\gamma_x)\sum_{t=1}^{\tau} \hat{\ell}^x_t(a) - \sum_{t=1}^{\tau}\sum_{a} q^x_t(i)\hat{\ell}^x_t(a)} \leq {\frac{\log K}{\eta_x} + \eta_x \sum_{t=1}^{\tau}\sum_{i=1}^{K} {q}^x_t(i)(\hat{\ell}^x_t(i))^2}
    \label{eq:exp3gurantee}
\end{align}

\textbf{Step 2:} Since the LHS in \eqref{eq:exp3gurantee} holds for every $a \in [K]$ we relate it to the regret $\widetilde{\optbx} - \sum_{t=1}^{\tau}\tilde{b}(x_t)$ using the following lemma.
\begin{restatable}{lemm}{lemmatermone}
\label{lemma:lhs} For any $a \in [K]$
        \begin{align*}
            (1&-\gamma_x)\sum_{t=1}^{\tau} \hat{\ell}^x_t(a) - \sum_{t=1}^{\tau}\sum_{a} q^x_t(i)\hat{\ell}^x_t(a) \geq \widetilde{\optbx} - \sum_{t=1}^{\tau}\tilde{b}(x_t) -\cO(Z+1)\sqrt{T \log T} - \gamma_x (Z_x+1)
        \end{align*}
\end{restatable}
Next we upper bound the RHS in \eqref{eq:exp3gurantee} using the following lemma.
\begin{restatable}{lemm}{lemmatermtwo}
\label{lemma:rhs}
         For $\eta_x = \big(\frac{\log K}{T\sqrt{K}}\big)^{2/3}\frac{1}{2Z_x + 1}$ and $\gamma_x = \sqrt{\eta_x K}$
        \begin{align*}
            \frac{\log K}{\eta_x} &+ \eta_x \sum_{t=1}^{\tau}\sum_{i=1}^{K} {q}^x_t(i)(\hat{\ell}^x_t(i))^2 \leq \cO\left(\Big(\frac{\optb_x}{B}+1\Big)(K \log K)^{1/3}T^{2/3}\right)
        \end{align*}
\end{restatable}

\textbf{Step 3:} We repeat the same argument for the second arm choice and then combining with Lemma~\ref{lemm:lemmaoptshifted} and Lemma~\ref{lemm:lemmaoptmarginal} completes the proof.

\section{Experiments}
\label{sec:exp}

We test our proposed algorithm \vig on both synthetic and real world data in the constrained setting against existing Borda dueling bandit algorithms that do not factor in the associated consumptions. We briefly describe our datasets, benchmarks and results here (also see Appendix~\ref{sec:exp_app}).

\textbf{Datasets.} We run our experiments on two datasets.
\begin{enumerate}
\item\textbf{Synthetic Data:} We create a Constrained Dueling Bandits instance with $K = 6$       arms (see Appendix~\ref{sec:exp_app} for the exact description of the preference matrix).
    The vector of Borda scores $\Bar{b} = \begin{pmatrix}
    b(1)& b(2)6& \ldots& b(6)
    \end{pmatrix}^{\top}$ is given by
    $\begin{pmatrix}
    0.672& 0.646& 0.602& 0.582& 0.554& 0.544
    \end{pmatrix}^{\top}.$
    We experiment with three choices of consumptions. In all three cases the number of resources $d=1$ and the true consumptions across both arms choices are given by the same function, i.e., $u^* = v^*$, and we add zero mean gaussian noise to each entry. The vector of consumptions for arms $\bar{u}^* =\begin{pmatrix}
    u^*(1)& u^*(2)6& \ldots& u^*(6)
    \end{pmatrix}^{\top}$ are given by $\begin{pmatrix}
    0.9&0.9&0.1&0.8&0.8&0.8
    \end{pmatrix}^{\top}$, $\begin{pmatrix}
    0.1&0.2&0.3&0.4&0.5&0.6
    \end{pmatrix}^{\top},$ and $\begin{pmatrix}
    0&0&0&0&0&0
    \end{pmatrix}^{\top}$.
    In the first case although arm 1 and 2 have high Borda scores, the associated consumptions are also high. In the second case the order of consumptions is the same as the order of Borda scores. In the last case all the consumptions are zero and our objective is to evaluate if our algorithm under performs in the absence of constraints.
    The experiments are run for $T = 2000$ rounds with $B=1000$ and are run independently over 50 samples.
    
\item\textbf{Car preference dataset:} We consider the Car preference dataset from \cite{Abbasnejad2013} where the preference matrix is generated by considering the user preferences for various models of cars. As in case 1, we consider three choices of consumptions that follow a similar structure(see Appendix~\ref{sec:exp_app} for more details).
The experiments are run for $T = 5000$ rounds with $B=4000$ and are run independently over 50 samples.

\end{enumerate}
\textbf{Benchmarks.} We compare our algorithm against the following two choices of DB algorithms for Borda scores. 
\begin{enumerate}
\item \textbf{D-EXP3}: Dueling EXP3 algorithm (Algorithm 1, \cite{pmlr-v139-saha21a}) runs an exponential weights algorithm with uniform exploration on the estimated Borda scores. 

\item \textbf{D-TS:} Dueling Thompson Sampling (Algorithm 2, \cite{Lekang2019SimpleAF}) runs a Thompson sampling algorithm by learning true parameter values, which can represent the preference matrix directly or by some other latent values for each action, by sampling the posterior distribution conditioned on the history. As in \cite{Lekang2019SimpleAF}, we use $K^2 - K$ independent Beta(1, 1) as our prior.
\end{enumerate}

\begin{figure*}[t!]
    \centering
    \subfigure{\includegraphics[width=0.32\textwidth]{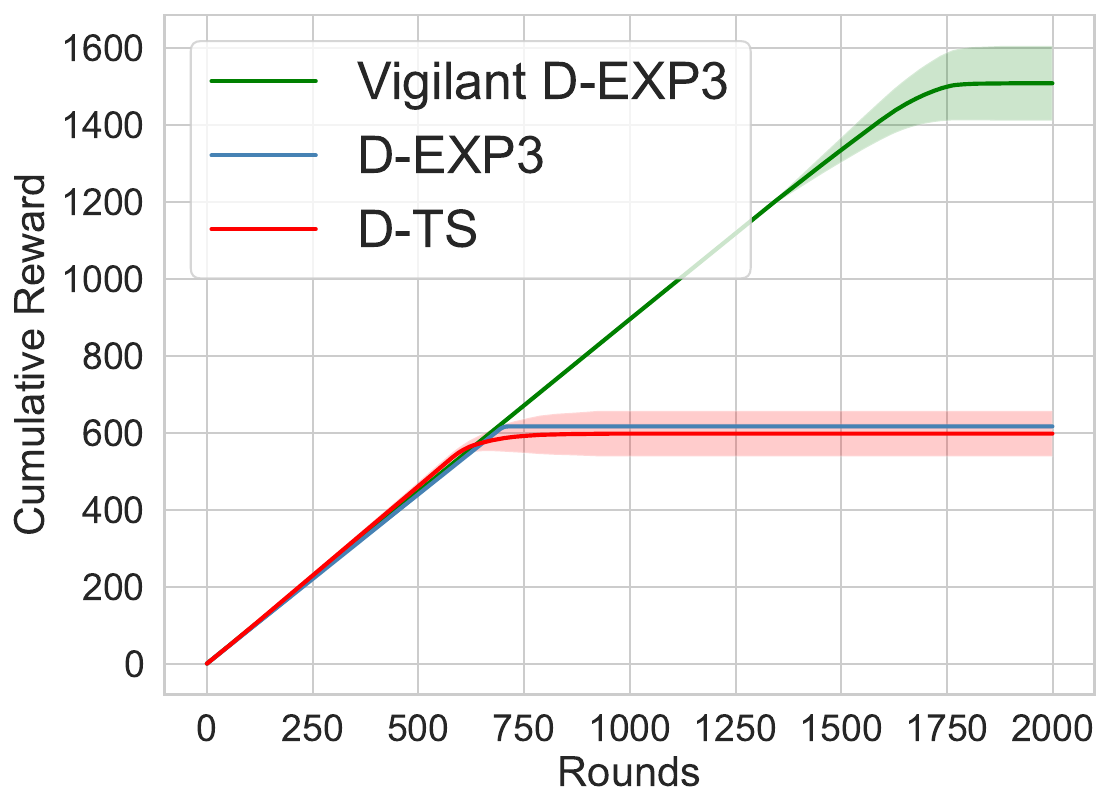}} 
    \subfigure{\includegraphics[width=0.32\textwidth]{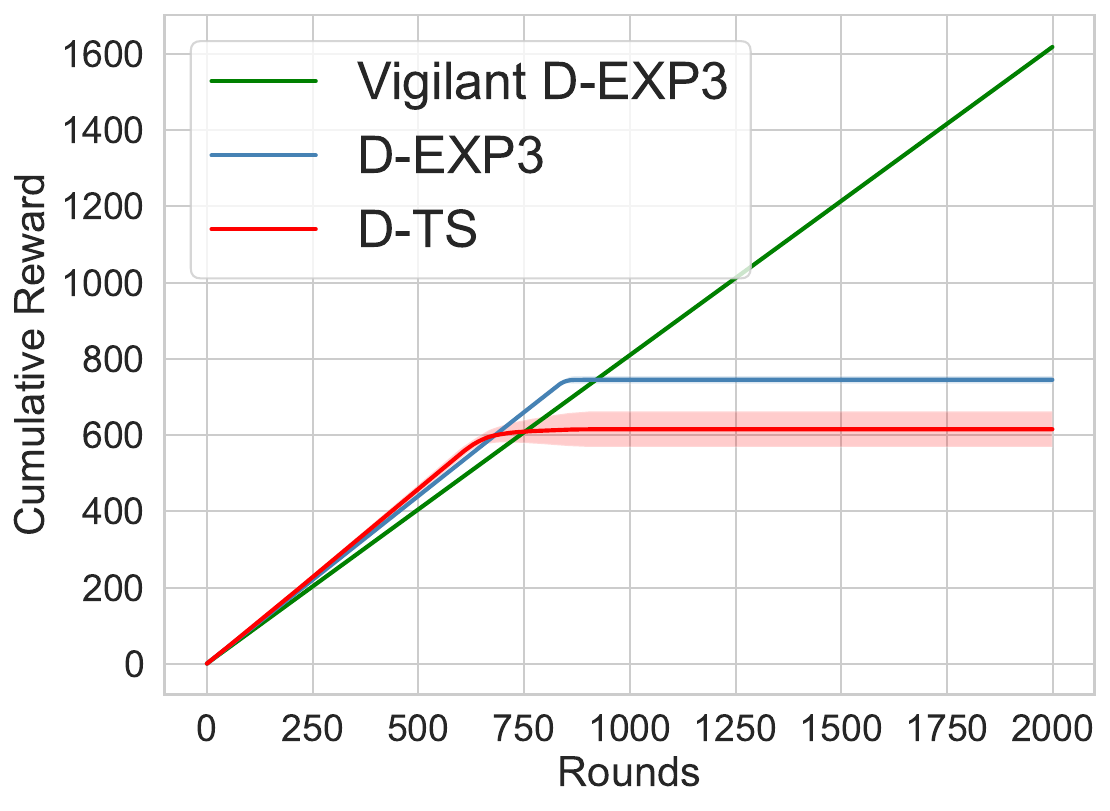}} 
    \subfigure{\includegraphics[width=0.32\textwidth]{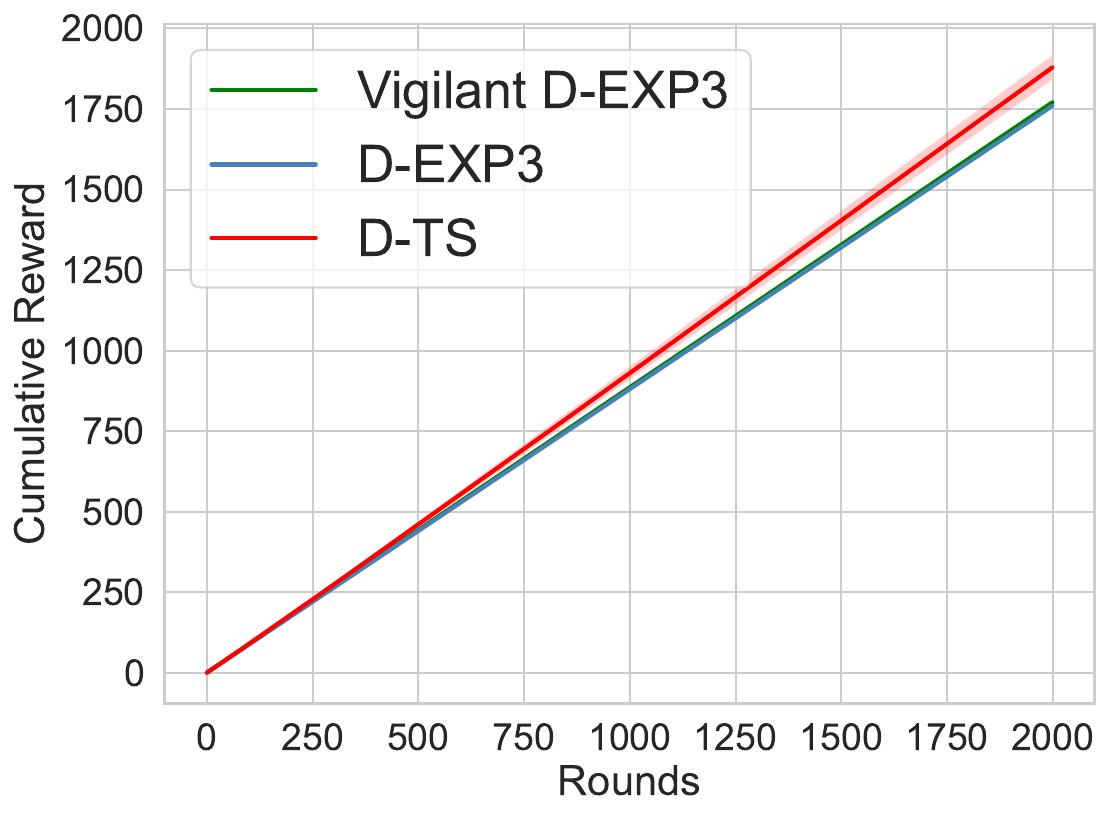}}
    \caption{Cumulative Reward across Rounds on synthetic data for three choices of consumptions. The first corresponds to the case when the consumption of an arm with intermediate Borda score is lowest, second to when the order of consumptions is the same as the order of Borda score and third corresponds to zero consumption. } 
    \label{fig:1}
\end{figure*}

\begin{figure*}[t!]
    \centering
    \subfigure{\includegraphics[width=0.32\textwidth]{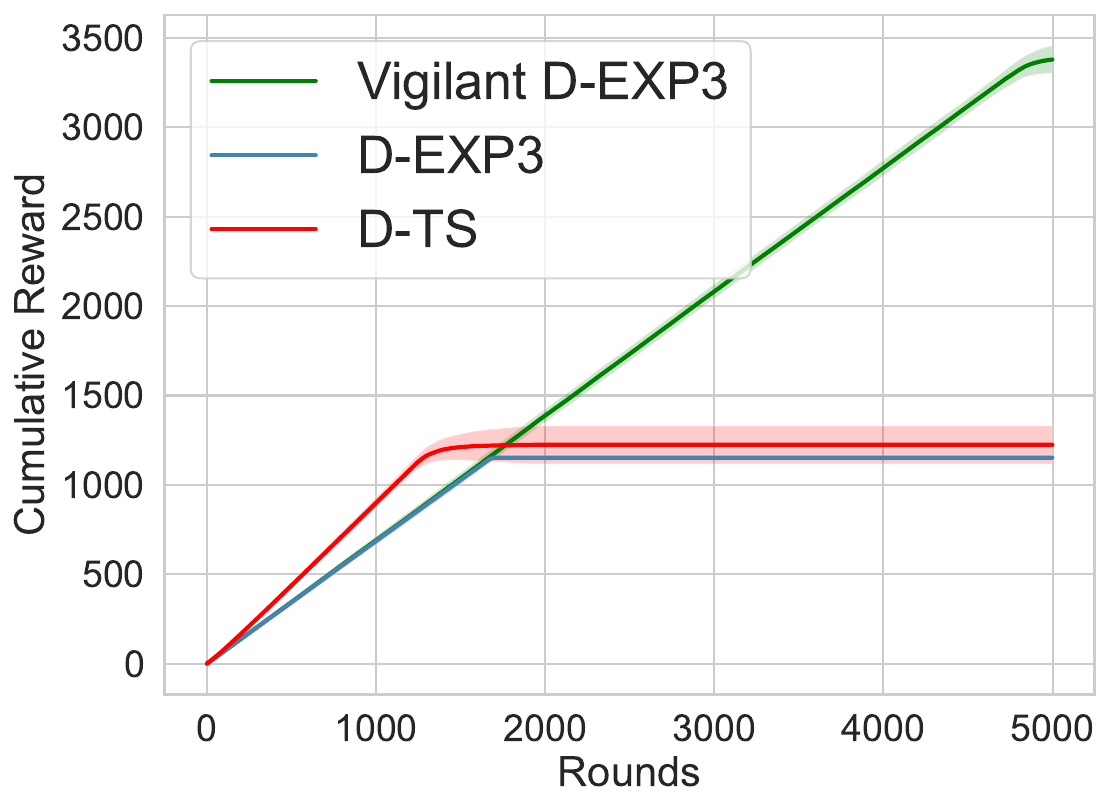}} 
    \subfigure{\includegraphics[width=0.32\textwidth]{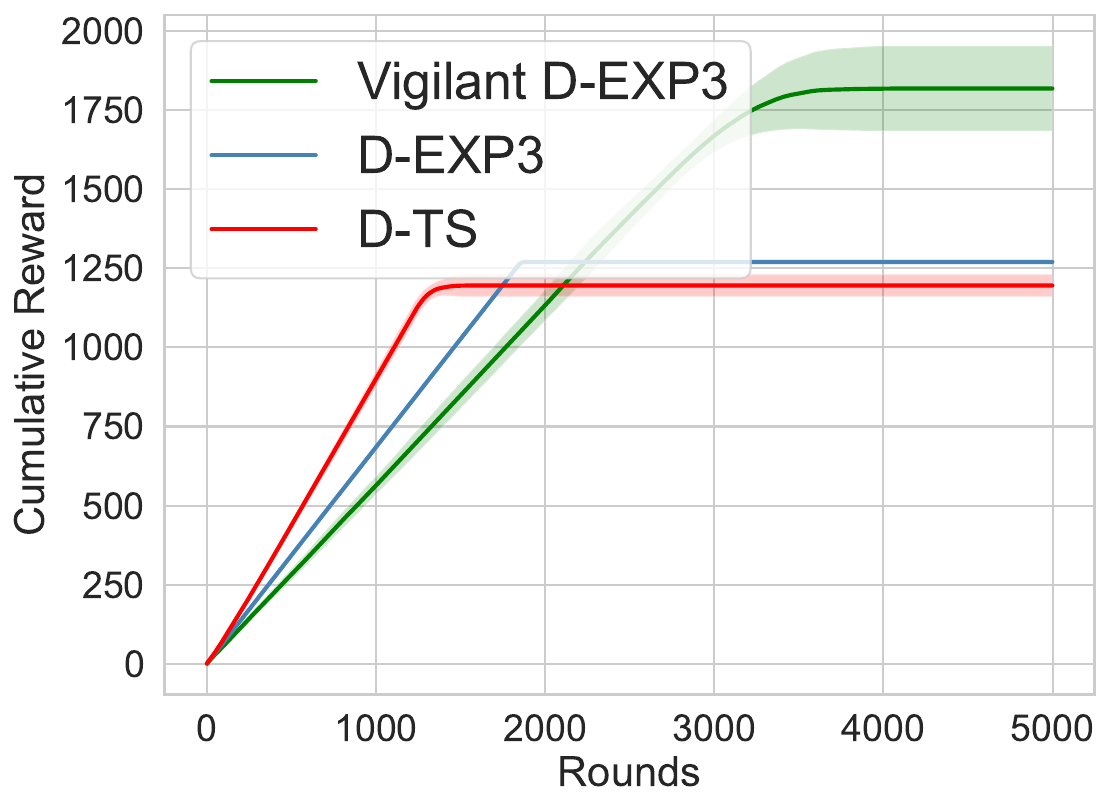}} 
    \subfigure{\includegraphics[width=0.32\textwidth]{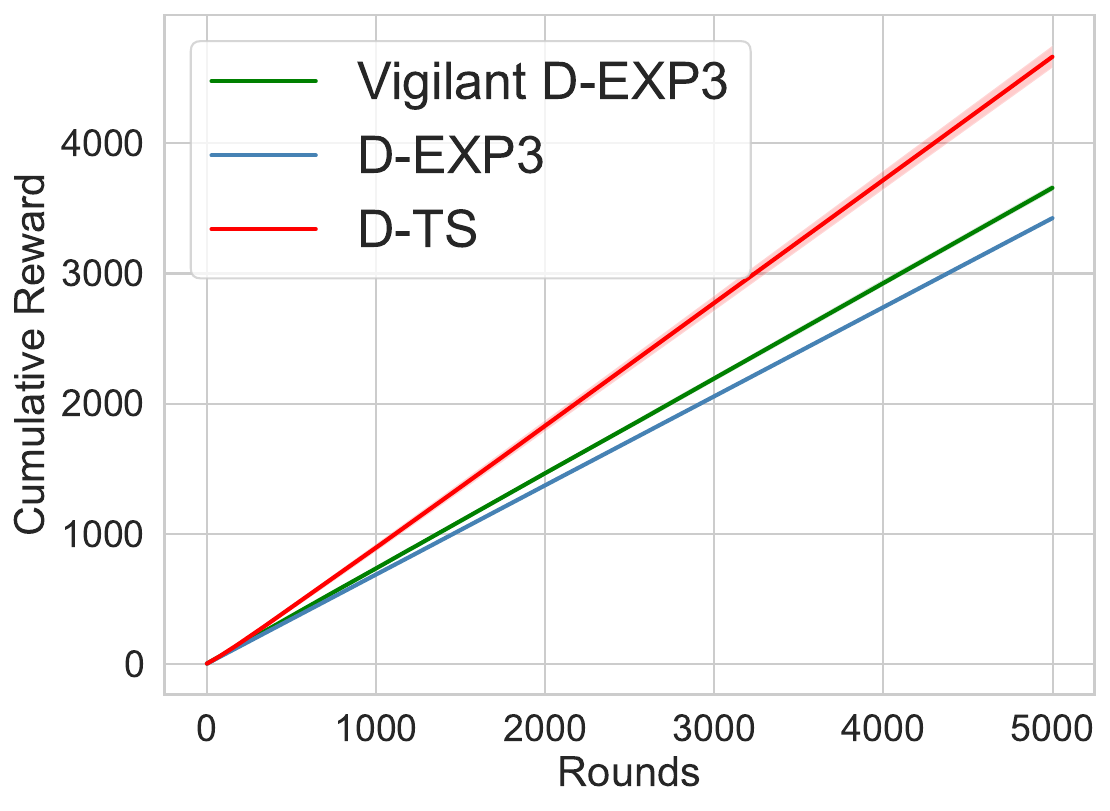}}
    \caption{Cumulative Reward across Rounds on car preference dataset for three choices of consumptions. The first corresponds to the case when the consumption of an arm with intermediate Borda score is lowest, second to when the order of consumptions is the same as the order of Borda score and third corresponds to zero consumption. } 
    \label{fig:2}
\end{figure*}

\textbf{Results.}
Figure~\ref{fig:1} and Figure~\ref{fig:2} plot the cumulative rewards across different rounds. For both datasets in the first two cases when it is not prudent to stick to the arms that have high Borda scores since they also have high associated consumptions, the benchmark algorithms earn more reward in the initial few rounds but stop early since they run out of budget. In contrast, our algorithm \vig, does take into account the associated consumptions and runs for far more number of rounds before exhausting its resources and therefore ends up acquiring far more reward. However in the unconstrained case although the performance of \vig\; is almost same as $\mathtt{D\text{-}EXP3}$, $\mathtt{D\text{-}TS}$ outperforms them both and as such an immediate direction of study would be to develop a constrained version of Dueling Thompson Sampling and compare its performance against \vig.


\section{Conclusion}
\label{sec:conclusion}
In this work we have developed the framework for learning from preference feedback under resource constraints and developed several lower bounds both with Borda scores and Condorcet scores that show that the setting is strictly more difficult than its multi-armed counterpart. Under the assumption that the resource budget is high, we developed an EXP3 based algorithm that via the lagrangian also takes into account the associated consumptions of a duel rather than just the associated scores. We show that the algorithm enjoys sub-linear regret bound and performs far better on both synthetic and real world data. 

The Condorcet constrained DB problem is a strictly harder problem, since to be able to compute an estimate of the Condorcet score one needs to know the identity of the Condorcet winner. 
Further none of the elimination based algorithms for Condorcet winner can be used here since if the winner is eliminated based say on the lagrangian value, then further estimates of the scores cannot be computed and as such makes the problem quite challenging and developing algorithms for this setting is an interesting future work.
Moreover as observed in Section~\ref{sec:exp} Dueling Thompson sampling appears to perform better in the unconstrained setting and as such it might be instructive to develop a constrained version of the algorithm. Finally, it would certainly be useful to consider more general and practical settings of dueling bandits. 

\newpage

\bibliographystyle{abbrvnat} 
\bibliography{references,dbrefs,assort}

\newpage
\appendix
\section{Extended Related Works}
\label{sec:App_related}
In Section~\ref{sec:related_works} we provided an overview of some existing works in Dueling Bandits and Canstrained Bandits. Here we provide a more comprehensive study of these two areas.

\begin{itemize}
    \item \textbf{Dueling Bandits:}
    
    Over the past decade, the Dueling Bandits setting has undergone significant advancements. The problem, in its current form, was introduced by Yue and Joachims (2012) \cite{YUE20121538}, who established upper and lower bounds on regret. These bounds were established under the assumption that the preference matrix has specific structures, such as total ordering, strong stochastic transitivity, and strong triangle inequality (refer to Section~\ref{sec:prelim} for definitions).
    
    Subsequently \citep{btm} proposed `Beat the Mean' algorithm with improved regret bound while also relaxed the strong stochastic transitivity assumption to relaxed stochastic transitivity. \citet{rucb} further relaxed the total ordering assumption to the existence of a Condorcet winner (an arm that beats every other arm) and provided a upper confidence bound (UCB) based algorithm. 
    
    \cite{Zoghi2015b} examine the same problem, but emphasizes learning situations where a vast number of arms are available. To minimize the number of comparisons, the authors introduce the MergeRUCB algorithm that employs a divide-and-conquer strategy akin to the merge sort algorithm. It begins by organizing the arms into predefined batches, processing each batch independently before combining the results. \cite{Zoghicope,pmlr-v48-komiyama16} study the dueling bandits problem focusing on Copeland winners and \cite{articleli} suggest a thompson sampling based algorithm to solve it.
    
    \cite{ailon} studied the dueling bandit problem in an adversarial setup (where the preference matrix $P$ changes over time), introducing the sparring EXP3 idea, albeit without regret guarantees. Subsequent works~\cite{gajane15, pmlr-v139-saha21a, pmlr-v162-saha22b} utilized this concept to prove regret guarantees in adversarial environments. \cite{pmlr-v139-saha21a} consider the Borda winner instead of Condorcet winner, which is known to always exist. \cite{Lekang2019SimpleAF} provided both Thompson sampling and sparring EXP3 type algorithm for both maxmin and Borda winners along with regret guarantees and show impressive improvement on performance against existing benchmark algorithms. 

    Over the last two decades the field of Dueling Bandits received significant attention due to the simplicity and effectiveness of the problem framework. 
    Consequently Dueling Bandits has been generalized and studied for various real world problems, including but not limited to, extending pairwise preferences to subsetwise preferences \cite{SG19,SGrank18,Sui+17,SGho21}, large decision spaces \cite{S21,SKM22}, adversarial preferences \cite{gupta2021optimal} and contextual scenarios
\cite{SK21,balsubramani16}, item unavailability \cite{SG21dbaa}, non-stationary preferences \cite{buening2023anaconda}. Consequently, the framework of Dueling Bandit has also been adapted to other interdisciplinary fields of research, e.g., reinforcement learning \cite{SPL21,BlumOpt}, robotics \cite{lee2021pebble,rlhf}, language models \cite{chatgpt,instructgpt} and assortment optimization \cite{assort-mnl,assort-mallows}.
A detailed survey of Preference Bandit literature could be found in \citep{survey_duel,sui2018advancements}.

    \item \textbf{Constrained Bandits:}

    There is a body of literature that under the name \emph{Bandits with Knapsacks} that looks at cumulative reward maximization under budget constraints. It was first introduced in \citep{Badanidiyuru_2013} for the MAB setting and the proposed algorithms- BalancedExploration and PrimalDualBwK were shown to enjoy optimal regret bounds up to polylogarithmic factors. BalancedExploration however is not an efficient algorithm (see Remark 4.2 in \citep{Badanidiyuru_2013}) and we do not pursue this. We further show that in fact PrimalDualBwK attains an $\Omega(T)$ regret in the MAB setting, and as such we do not try to adapt this algorithm to the Constrained-DB setting. 
    
    Subsequently, more general versions of the problem have been studied, eg., in the linear contextual setting, \citep{linCBwK} provide an algorithm utilizing ideas from UCB and primal-dual methods with sub-linear regret bound. The idea is to maintain a UCB estimate of the associated lagrangian by maintaining a UCB estimate of the rewards and an LCB estimate of the consumptions and the arm being played is the one that maximized the lagrangian optimistically, while the lagrange multiplier is updated via a dual optimization step. In the fully adversarial setting,\citep{Immorlica} show that regret minimization is not feasible and therefore provide gurantees on the competitive ratio of the proposed algorithm. In the smooth adversarial setting, where in the contexts and rewards are chosen by an adaptive adversary but nature perturbs it using a small gaussian noise, \citep{pmlr-v162-sivakumar22a} provide sub-linear regret again using a primal-dual idea from \citep{linCBwK}.
    
    More general versions, where only realizability of the reward and consumptions by a function class is assumed have been studied in \cite{slivkins2023contextual,pmlr-v206-han23b} where the Inverse gap weighting idea from \cite{abe1999associative} has been employed to provide sub-linear regret bounds by computationally efficient algorithms. Finally recent works have also started study regret guarantees of algorithms in the reinforcement learning setting, where instead of a single state, the agent interacts with the environment through a sequence of states and actions \cite{pmlr-v130-ding21d,NEURIPS2022_14a5ebc9,Kalagarla_Jain_Nuzzo_2021,Wei_Liu_Ying_2022}.
\end{itemize}

\section{Proof of Lower Bounds}
\label{sec:App_regret_lower}
\subsection{Lower Bounds for Condorcet Constrained-DB}
\lemmaCWLBone*
\begin{proof}
    \textbf{General setting:}  
    We start with the general setting without any assumption on the preference matrix and subsequently consider total ordering and strong stochastic transitivity. The proof relies on creating $K-1$ problem instances that we denote by $\cI_{1},\ldots,\cI_{K-1}$.

    $\cI_{1}$ is defined by the following preference matrix:
    \begin{align*}
    P &= 
    \begin{bmatrix}
        \dfrac{1}{2} & \dfrac{1}{2} + \epsilon & \dfrac{1}{2} + 2\epsilon &\cdots & \dfrac{1}{2} + (K-1)\epsilon\\[6pt]
        \dfrac{1}{2} - \epsilon & \dfrac{1}{2} & \dfrac{1}{2}&\cdots&\dfrac{1}{2}\\[6pt]
        \dfrac{1}{2} - 2\epsilon & \dfrac{1}{2} & \dfrac{1}{2}&\cdots&\dfrac{1}{2}\\[6pt]
        \vdots & \vdots & \vdots & \ddots & \vdots\\[6pt]
        \frac{1}{2} - (K-1)\epsilon &  \dfrac{1}{2} & \dfrac{1}{2} &\cdots& \dfrac{1}{2}
    \end{bmatrix} 
\end{align*}
with $ u^*(1) = v^*(1) = 1, u^*(i) = v^*(i)= 0, \; \forall i \in \{2,\ldots,K\}$.

$\cI_{2}$ is defined by the following preference matrix:
    \begin{align*}
    P &= 
    \begin{bmatrix}
        \dfrac{1}{2} & \dfrac{1}{2} + (K-1)\epsilon & \dfrac{1}{2} + \epsilon &\dfrac{1}{2} + 2\epsilon &\cdots & \dfrac{1}{2} + (K-2)\epsilon\\[6pt]
        \dfrac{1}{2} - (K-1)\epsilon & \dfrac{1}{2} & \dfrac{1}{2}& \dfrac{1}{2}& \cdots&\dfrac{1}{2}\\[6pt]
        \dfrac{1}{2} - \epsilon & \dfrac{1}{2} & \dfrac{1}{2}&\dfrac{1}{2}&\cdots&\dfrac{1}{2}\\[6pt]
        \dfrac{1}{2} - 2\epsilon & \dfrac{1}{2} & \dfrac{1}{2}&\dfrac{1}{2}&\cdots&\dfrac{1}{2}\\[6pt]
        \vdots & \vdots & \vdots &\vdots& \ddots & \vdots\\[6pt]
        \frac{1}{2} - (K-2)\epsilon &  \dfrac{1}{2} & \dfrac{1}{2} & \dfrac{1}{2} &\cdots& \dfrac{1}{2}
    \end{bmatrix} 
\end{align*}
with $ u^*(1) = v^*(1) = 1, u^*(i) = v^*(i)= 0, \; \forall i \in \{2,\ldots,K\}$ and so on with
$\cI_{K-1}$ is defined by the following preference matrix:
    \begin{align*}
    P &= 
    \begin{bmatrix}
        \dfrac{1}{2} & \dfrac{1}{2} + 3\epsilon  &\cdots & \dfrac{1}{2} + \epsilon & \dfrac{1}{2} + 2\epsilon\\[6pt]
        \dfrac{1}{2} - 3\epsilon & \dfrac{1}{2} & \cdots& \dfrac{1}{2}&\dfrac{1}{2}\\[6pt]
        \vdots & \vdots & \vdots &\vdots& \vdots\\[6pt]
        \dfrac{1}{2} - \epsilon  & \dfrac{1}{2}&\dfrac{1}{2}&\cdots&\dfrac{1}{2}\\[6pt]
        \dfrac{1}{2} - 2\epsilon & \dfrac{1}{2}&\dfrac{1}{2}&\cdots&\dfrac{1}{2}\\[6pt]
    \end{bmatrix} 
\end{align*}
with $ u^*(1) = v^*(1) = 1, u^*(i) = v^*(i)= 0, \; \forall i \in \{2,\ldots,K\}$.

The optimal policy in $\cI_k$ plays arms $x_t = 1, y_t = 1$ for $B$ rounds and thereafter plays arms $x_t = k+1, y_t = k+1$ for the remaining $T-B$ rounds. The total accumulated reward by the optimal policy is $\optc = B + (T-B)\Big(\frac{1}{2} - \epsilon\Big)$. 

Note that arm $k+1$ is the Condorcet winner in $\cI_k$ and any algorithm needs to play the pairs $(1,k+1)$ and $(1,k+2)$ at least $\frac{1}{\epsilon^2}$ number of times to determine if $c(k+1) > c(k+2)$.  To differentiate between the $K-1$ instances we need at least $\Theta(\frac{K}{\epsilon^2})$ comparisons of arm 1 with $j\in \{2,\ldots,K\}$. However since the budget $B = o\left(\frac{K}{\epsilon^2}\right)$, no algorithm can differentiate between these instances and hence would always end up playing the sub-optimal arm at least $\frac{(T-B)}{2}$ number of times after the initial $B$ rounds. Therefore 
\begin{align*}
    \rewc \leq B + \frac{(T-B)}{2}\Big(\frac{1}{2} - \epsilon\Big) + \frac{(T-B)}{2}\Big(\frac{1}{2} - 2\epsilon\Big)
\end{align*}
which implies $\optc \geq \frac{(T-B)}{2}\epsilon = \Omega(T)$

\textbf{Total Ordering:} Next consider the following instances:

$\cI_{1}$ is defined by the following preference matrix:
    \begin{align*}
    P &= 
    \begin{bmatrix}
        \dfrac{1}{2} & \dfrac{1}{2} + \epsilon & \dfrac{1}{2} + 2\epsilon &\cdots & \dfrac{1}{2} + (K-1)\epsilon\\[6pt]
        \dfrac{1}{2} - \epsilon & \dfrac{1}{2} & \dfrac{1}{2} + \epsilon &\cdots&\dfrac{1}{2} + (K-2)\epsilon\\[6pt]
        \dfrac{1}{2} - 2\epsilon & \dfrac{1}{2} -\epsilon & \dfrac{1}{2}&\cdots&\dfrac{1}{2} + (K-3)\epsilon\\[6pt]
        \vdots & \vdots & \vdots & \ddots & \vdots\\[6pt]
        \frac{1}{2} - (K-1)\epsilon &  \dfrac{1}{2} - (K-2)\epsilon & \dfrac{1}{2} - (K-3)\epsilon &\cdots& \dfrac{1}{2}
    \end{bmatrix} 
\end{align*}
with $ u^*(1) = v^*(1) = 1, u^*(i) = v^*(i)= 0, \; \forall i \in \{2,\ldots,K\}$.

$\cI_{2}$ is defined by the following preference matrix:
    \begin{align*}
    P &= 
    \begin{bmatrix}
        \dfrac{1}{2} & \dfrac{1}{2} + (K-1)\epsilon & \dfrac{1}{2} + \epsilon &\dfrac{1}{2} + 2\epsilon &\cdots & \dfrac{1}{2} + (K-2)\epsilon\\[6pt]
        \dfrac{1}{2} - (K-1)\epsilon & \dfrac{1}{2} & \dfrac{1}{2} + (K-2)\epsilon& \dfrac{1}{2} + \epsilon& \cdots&\dfrac{1}{2} + (K-3)\epsilon\\[6pt]
        \dfrac{1}{2} - \epsilon & \dfrac{1}{2}  - (K-2)\epsilon & \dfrac{1}{2}&\vdots&\cdots&\dfrac{1}{2}  + (K-4)\epsilon\\[6pt]
        \dfrac{1}{2} - 2\epsilon & \dfrac{1}{2} - \epsilon& \vdots&\dfrac{1}{2}&\cdots&\vdots\\[6pt]
        \vdots & \vdots & \vdots &\vdots& \ddots & \vdots\\[6pt]
        \frac{1}{2} - (K-2)\epsilon &  \dfrac{1}{2} - (K-3)\epsilon & \dfrac{1}{2}  - (K-4)\epsilon & \cdots &\cdots& \dfrac{1}{2}
    \end{bmatrix} 
\end{align*}
with $ u^*(1) = v^*(1) = 1, u^*(i) = v^*(i)= 0, \; \forall i \in \{2,\ldots,K\}$ and so on. Corresponding to each of these instances we define the parallel set of instances given by: 

$\cI'_{1}$ is defined by the following preference matrix:
    \begin{align*}
    P' &= 
    \begin{bmatrix}
        \dfrac{1}{2} & \dfrac{1}{2} + \epsilon & \dfrac{1}{2} + 2\epsilon &\cdots & \dfrac{1}{2} + (K-1)\epsilon\\[6pt]
        \dfrac{1}{2} - \epsilon & \dfrac{1}{2} & \dfrac{1}{2} - \epsilon &\cdots&\dfrac{1}{2} - (K-2)\epsilon\\[6pt]
        \dfrac{1}{2} - 2\epsilon & \dfrac{1}{2} + \epsilon & \dfrac{1}{2}&\cdots&\dfrac{1}{2} - (K-3)\epsilon\\[6pt]
        \vdots & \vdots & \vdots & \ddots & \vdots\\[6pt]
        \frac{1}{2} - (K-1)\epsilon &  \dfrac{1}{2} + (K-2)\epsilon & \dfrac{1}{2} + (K-3)\epsilon &\cdots& \dfrac{1}{2}
    \end{bmatrix} 
\end{align*}
with $ u^*(1) = v^*(1) = 1, u^*(i) = v^*(i)= 0, \; \forall i \in \{2,\ldots,K\}$.

$\cI'_{2}$ is defined by the following preference matrix:
    \begin{align*}
    P' &= 
    \begin{bmatrix}
        \dfrac{1}{2} & \dfrac{1}{2} + (K-1)\epsilon & \dfrac{1}{2} + \epsilon &\dfrac{1}{2} + 2\epsilon &\cdots & \dfrac{1}{2} + (K-2)\epsilon\\[6pt]
        \dfrac{1}{2} - (K-1)\epsilon & \dfrac{1}{2} & \dfrac{1}{2} + (K-2)\epsilon& \dfrac{1}{2} + \epsilon& \cdots&\dfrac{1}{2} - (K-3)\epsilon\\[6pt]
        \dfrac{1}{2} - \epsilon & \dfrac{1}{2}  - (K-2)\epsilon & \dfrac{1}{2}&\vdots&\cdots&\dfrac{1}{2}  - (K-4)\epsilon\\[6pt]
        \dfrac{1}{2} - 2\epsilon & \dfrac{1}{2} - \epsilon& \vdots&\dfrac{1}{2}&\cdots&\vdots\\[6pt]
        \vdots & \vdots & \vdots &\vdots& \ddots & \vdots\\[6pt]
        \frac{1}{2} - (K-2)\epsilon &  \dfrac{1}{2} + (K-3)\epsilon & \dfrac{1}{2}  + (K-4)\epsilon & \cdots &\cdots& \dfrac{1}{2}
    \end{bmatrix} 
\end{align*}
with $ u^*(1) = v^*(1) = 1, u^*(i) = v^*(i)= 0, \; \forall i \in \{2,\ldots,K\}$ and so on. 

Notice that although both set of instances $\cI_1,\ldots,\cI_K$ and $\cI'_1,\ldots,\cI'_K$ have total ordering, the total ordering in $\cI'_1,\ldots,\cI'_K$ does not match the ordering of the Condorcet scores in $\cI'_1,\ldots,\cI'_K$. Therefore dueling the sub-optimal arms $\{2,\ldots,K\}$ does not give any information about the order of Condorcet scores.

\textbf{Strong Stochastic Transitivity:} Finally suppose we assume that the Preference matrix follows strong stochastic transitivity, i.e., if $i \succ j$ in the total ordering sense then $P(i,k) \geq \max\{P(i,j),P(j,k)\}$. With this assumption, notice that the instances $\cI'_1,\ldots,\cI'_K$ are not allowed and therefore the algorithm may learn about the Condorcet order from the total order relations.
\end{proof}

\lemmaCWLBtwo*
\begin{proof}
    We provide the proof for $K=3$; the general case can be proven by constructing $K-1$ such instances as in the proof of Lemma~\ref{lemma:condorcet_lower_bound_2}. Consider the following preference matrix:
    \begin{align*}
    P' &= 
    \begin{bmatrix}
        \dfrac{1}{2} & \dfrac{1}{2} + \epsilon & \dfrac{1}{2} + 2\epsilon\\[6pt]
        \dfrac{1}{2} - \epsilon & \dfrac{1}{2} & \dfrac{1}{2}+\epsilon\\[6pt]
        \dfrac{1}{2} - 2\epsilon & \dfrac{1}{2}-\epsilon & \dfrac{1}{2}
    \end{bmatrix} 
    \end{align*}
    with the following consumptions: $u^*(1) = v^*(1) = 1, u^*(2) = v^*(2) = \epsilon$ and $u^*(3) = v^*(3) = 0$. The proof then follows as in the proof of Lemma~\ref{lemma:bbbborda_lower_bound_2}.
\end{proof}

\subsection{Lower Bounds for Borda Constrained-DB}
\lemmaBWLBone*
\begin{proof}
    We provide the proof for $K=3$; the general case can be proven by constructing $K-1$ such instances as in the proof of Lemma~\ref{lemma:condorcet_lower_bound_2}. Consider the following preference matrix:
    \begin{align*}
    P &= 
        \begin{bmatrix}
            \dfrac{1}{2} & \dfrac{1}{2} & \dfrac{1}{2} + 2\epsilon\\[6pt]
            \dfrac{1}{2} & \dfrac{1}{2} & \dfrac{1}{2}+\epsilon\\[6pt]
            \dfrac{1}{2} - 2\epsilon & \dfrac{1}{2}-\epsilon & \dfrac{1}{2}
        \end{bmatrix} 
    \end{align*}
    with the following consumptions: $u^*(1) = v^*(1) = 0, u^*(2) = v^*(2) = 0$ and $u^*(3) = v^*(3) = 1$. Note that the Borda scores are given by $b(1) = 1/2 + \epsilon, b(2) = 1/2 + \epsilon/2$ and $b(3) = 1/2 - 3\epsilon/2$. The Borda winner is arm 1 and the gap between second best arm and the Borda winner is $\Theta(\epsilon)$. Also note that the optimal policy always plays arm 1. To be able to differentiate between them we need to play $(1,2)$ and $(1,3)$ at least $\cO(\frac{K}{\epsilon^2})$. Since $B = o(\frac{K}{\epsilon^2})$, no algorithm can differentiate between arm 1 and arm 2 and therefore the regret of any algorithm will be $\Omega(T)$.
\end{proof}
\lemmaBWLBtwo*
\begin{proof}
    We provide the proof for $K=3$; the general case can be proven by constructing $K-1$ such instances as in the proof of Lemma~\ref{lemma:condorcet_lower_bound_2}. Consider the following preference matrix:
    \begin{align*}
    P &= 
        \begin{bmatrix}
            \dfrac{1}{2} & \dfrac{1}{2} & \dfrac{1}{2} + 2\epsilon\\[6pt]
            \dfrac{1}{2} & \dfrac{1}{2} & \dfrac{1}{2}+\epsilon\\[6pt]
            \dfrac{1}{2} - 2\epsilon & \dfrac{1}{2}-\epsilon & \dfrac{1}{2}
        \end{bmatrix} 
    \end{align*}
    with the following consumptions: $u^*(1) = v^*(1) = 0, u^*(2) = v^*(2) = \epsilon$ and $u^*(3) = v^*(3) = 1$. The optimal policy chooses arm 1 always. However, the gap between arm 1 and arm 2 is $\Theta(\epsilon)$ and therefore arms $(1,2)$ and $(1,3)$ have to be played $\Theta(\frac{K}{\epsilon^2})$ number of times. However, playing arm $2$ consumes $\epsilon$ amount and therefore $\Theta(\frac{K}{\epsilon^2})\epsilon$ budget is needed to differentiate between arms $1$ and $2$. Since $B = o(\frac{K}{\epsilon})$, any algorithm would incur a regret of $\Omega(T)$.
\end{proof}

\section{Proof of Regret Bound}
\label{sec:App_regret_proof}
\lemmaoptshifted*
\begin{proof}
Observe that 
    \begin{align*}
    \widetilde{\pi}^{*(b)} &= \underset{\pi_x,\pi_y \in \Delta^{K}}{\argmax} \sum_{x\in [K]} \pi_x(x) \tilde{b}(x)  + \sum_{y\in [K]} \pi_y(y)\tilde{b}(y)  \\
    & \text{such that } \sum_{x,y \in [K]} \pi_x(x) u^*(x)  +  \pi_y(y) v^*(y)  \leq \frac{B}{T} \mathbf{1}\\
    &= \underset{\pi_x,\pi_y \in \Delta^{K}}{\argmax} \sum_{x\in [K]} \pi_x(x) \left(\frac{K-1}{K}{b}(x) + \frac{1}{2K}\right)  + \sum_{y\in [K]} \pi_y(y)\left(\frac{K-1}{K}{b}(y) + \frac{1}{2K}\right)  \\
    & \text{such that } \sum_{x,y \in [K]} \pi_x(x) u^*(x)  +  \pi_y(y) v^*(y)  \leq \frac{B}{T} \mathbf{1}\\
    &= \underset{\pi_x,\pi_y \in \Delta^{K}}{\argmax}  \sum_{x\in [K]} \pi_x(x) {b}(x)  + \sum_{y\in [K]} \pi_y(y){b}(y)   \\
    & \text{such that } \sum_{x,y \in [K]} \pi_x(x) u^*(x)  +  \pi_y(y) v^*(y)  \leq \frac{B}{T} \mathbf{1}\\
    & = {\pi}^{*(b)},
\end{align*}
i.e., the policy that solves both the LPs are same. Therefore
\begin{align*}
    \widetilde{\optb} &=  T\sum_{x,y \in [K]}\, \widetilde{\pi}^{*(b)}_{x}\, \tilde{b}(x) + \widetilde{\pi}^{*(b)}_{y}\, \tilde{b}(y)\\
    &= T \sum_{x,y \in [K]}\, \widetilde{\pi}^{*(b)}_{x}\, \left(\frac{K-1}{K}{b}(x) + \frac{1}{2K}\right) + \widetilde{\pi}^{*(b)}_{y}\, \left(\frac{K-1}{K}{b}(y) + \frac{1}{2K}\right)\\
    &= \frac{K-1}{K} \left( T\, \sum_{x,y \in [K]}\, {\pi}^{*(b)}_{x}(x)\, b(x) +  {\pi}^{*(b)}_{y}(y)\, b(y)\right) + \frac{T}{K}\\
    &= \frac{K-1}{K} \optb + \frac{T}{K}
\end{align*}
Further 
\begin{align*}
    \widetilde{\rewb} &= \sum_{t=1}^{\tau} \tilde{b}(x_t) + \tilde{b}(y_t)\\
    &= \sum_{t=1}^{\tau} \left(\frac{K-1}{K}{b}(x_t) + \frac{1}{2K}\right) + \left(\frac{K-1}{K}{b}(y_t) + \frac{1}{2K}\right)\\
    &=  \sum_{t=1}^{\tau} {b}(x_t) + {b}(y_t) + \frac{\tau}{K}\\
    &= \rewb + \frac{\tau}{K}
\end{align*}
Therefore 
\begin{align*}
    \widetilde{\optb} - \widetilde{\rewb} &= \frac{K-1}{K} \optb - \rewb + \frac{T-\tau}{K}\\
    &\geq \optb - \rewb
\end{align*}
where the last line follows because $\tau \leq T$.
\end{proof}
\lemmaoptmarginal*
\begin{proof}
    We define the solution to \eqref{eq:LP-Shifted-Borda-x} as 
    \begin{align*}
        \tilde{\pi}_x^{(b)} = \underset{\pi_x \in \Delta^{K}}{\argmax} &\sum_{x\in [K]} \pi_x(x) \tilde{b}(x)\nonumber\\
    \text{such that } &\sum_{x \in [K]} \pi_x(x) u^*(x) \leq \frac{B}{2T} \mathbf{1}
    \end{align*}
    and the solution to \eqref{eq:LP-Shifted-Borda-x} as
    \begin{align*}
        \tilde{\pi}_y^{(b)} = \underset{\pi_x \in \Delta^{K}}{\argmax} &\sum_{x\in [K]} \pi_x(x) \tilde{b}(x)\nonumber\\
    \text{such that } &\sum_{x \in [K]} \pi_x(x) u^*(x) \leq \frac{B}{2T} \mathbf{1}
    \end{align*}
Now consider the the following distribution $\hat{\pi}(x) = \frac{1}{2}\tilde{\pi}_x^{(b)} + \frac{1}{2}\tilde{\pi}_y^{(b)} $. Note that $\sum_{x,y \in [K]} \hat{\pi}(x) u^*(x)  +  \hat{\pi}(y) v^*(y)  \leq \frac{B}{2T} \mathbf{1} + \frac{B}{2T} \mathbf{1} = \frac{B}{T} \mathbf{1}$ and therefore $\hat{\pi}$ is a feasible solution to \eqref{eq:LP-Shifted-Borda} and therefore 
\begin{align*}
        \widetilde{\optbx} + \widetilde{\optby} \geq \widetilde{\optb}.
    \end{align*}
\end{proof}
\theoremmain*
\begin{proof}
    The proof of the theorem follows along the following three steps:
    \begin{enumerate}
        \item[]\textbf{Step-1:} Note that 
        \begin{align*}
            \mid \hat{\ell}^x_t(a) \mid &= \bigg | \; \hat{b}_t(i) + Z_x \lambda_{t}^{\top} \Big(\frac{B}{2T}\mathbbm{1} - \hat{u}_t(i)\Big) \; \bigg |\\
            &\leq |\hat{b}_t(i)| + Z_x \|\lambda_t\|_{1} \left(\frac{B}{2T} \|\mathbbm{1}\|_{\infty} + \|\hat{u}_t\|_{\infty} \right)\\
            &\leq 1 + Z_x (1 + 1)\\
            &= 1 + 2Z_x
        \end{align*}
 Therefore $\frac{1}{2Z_x + 1}\hat{\ell}^x_t(a) \leq 1$ and using the regret guarantee of Exponential Weights algorithm \cite{auer2002finite}, \cite[Chapter 11]{lattimore_szepesvári_2020} we get for all $a \in [K]$
\begin{align*}
        \sum_{t=1}^{\tau} \hat{\ell}^x_t(a) - \sum_{t=1}^{\tau}\sum_{i} \tilde{q}^x_t(i)\hat{\ell}^x_t(i)  \leq \frac{\log K}{\eta_x} + \eta_x \sum_{t=1}^{\tau}\sum_{i=1}^{K} \tilde{q}^x_t(i)(\hat{\ell}^x_t(i))^2
\end{align*}
Since $\displaystyle \tilde{q}^x_t(i) = \frac{q^x_t(i) - \frac{\gamma_x}{K}}{1-\gamma_x}$, we have
\begin{align}
    \forall a \in [K], \; \underbrace{(1-\gamma_x)\sum_{t=1}^{\tau} \hat{\ell}^x_t(a)}_{I} - \underbrace{\sum_{t=1}^{\tau}\sum_{i=1}^{K} q^x_t(i)\hat{\ell}^x_t(i)}_{II}
    \leq {\frac{\log K}{\eta_x} + \eta_x \sum_{t=1}^{\tau}\sum_{i=1}^{K} {q}^x_t(i)(\hat{\ell}^x_t(i))^2}
    \label{eq:exp3gurantee_app}
\end{align}
    \item[] \textbf{Step-2:} Next we relate the LHS of \eqref{eq:exp3gurantee_app} to the regret using the following lemma.
\lemmatermone*
\begin{proof}
Let $\tilde{\pi}_x^{*(b)}$ be the solution of \eqref{eq:LP-Shifted-Borda-x}. Next define $\cH_{t-1} = \sigma\big(\{x_i,y_i,o_i(x_i,y_i)\}_{i=1}^{t-1}\big)$ be the sigma algebra generated by $\{x_i,y_i,o_i(x_i,y_i)\}_{i=1}^{t-1}$ and $\E_{\cH_{t-1}}$ be the conditional expectation with respect to $\cH_{t-1}$. Consider term $I$ and observe that 
\begin{align*}
    (1-\gamma_x)\sum_{t=1}^{\tau} \sum_{a} &\mathbb{E}_{\cH_{t-1}} \; [\hat{\ell}^x_t(a) \tilde{\pi}_x^{*(b)}(a)]
    = (1-\gamma_x)\sum_{t=1}^{\tau}\sum_{a} \tilde{\pi}_x^{*(b)}(a) \mathbb{E}_{\cH_{t-1}} \left[\hat{b}_t(a) + Z_x\lambda_t^{x\top}\Big(\frac{B}{2T}\mathbbm{1} - \hat{u}^x_t(a) \Big)\right].
\end{align*}
Note that $\E[\hat{b}_t(a)] = \tilde{b}(a)$ (see Lemma 4 in \cite{pmlr-v139-saha21a}) and $\E[\hat{u}^x_t(a)] = u^*(a)$. Using Azuma-Hoeffding inequality with probability at least $1 - \cO(\frac{1}{T^2})$
\begin{align*}
    (1-\gamma_x)\sum_{t=1}^{\tau} \sum_{a} \; \hat{\ell}^x_t(a) \tilde{\pi}_x^{*(b)}(a) \geq (1-\gamma_x)\sum_{t=1}^{\tau}\sum_{a} \tilde{\pi}_x^{*(b)}(a) \mathbb{E}_{\cH_{t-1}} \left[\tilde{b}(a) + Z_x\lambda_t^{x\top}\Big(\frac{B}{2T}\mathbbm{1} - {u}^*(a) \Big)\right] - \sqrt{T \log T}
\end{align*}

Next observe that with $D_t = \tilde{b}(a) + Z_x\lambda_t^{x\top}\Big(\frac{B}{2T}\mathbbm{1} - {u}^*(a) \Big) - \frac{\widetilde{\optbx}}{T}$ is adapted to $\cH_t,$ $|D_t| \leq 2(Z_x+1)$ and $\mathbb{E}_{\cH_{t-1}}[D_t] \geq 0$ and therefore using Azuma-Hoeffding with probability at least $1-\cO(1/T^2)$
\begin{align*}
    \sum_{t=1}^{\tau} \sum_{a} \; \hat{\ell}^x_t(a) &= \sum_{t=1}^{\tau} \sum_{a} \; \hat{\ell}^x_t(a) \tilde{\pi}_x^{*(b)}(a) \geq \frac{\tau}{T} \widetilde{\optbx} - 2(Z_x+1)\sqrt{T \log T}
\end{align*}
Therefore 
\begin{align}
 \label{eq:lhs_termI}
     (1-\gamma_x)\sum_{t=1}^{\tau} \sum_{a} \hat{\ell}^x_t(a) \geq \frac{\tau}{T} \widetilde{\optbx} - \cO(Z+1)\sqrt{T \log T} - \gamma_x (Z_x+1)
\end{align}
Next consider term $II$ from \eqref{eq:exp3gurantee_app}:
\begin{align*}
    \sum_{t=1}^{\tau}\sum_{a} q^{x}_t(a)\hat{\ell}^{x}_t(a) &= \sum_{t=1}^{\tau}\sum_{a} q^x_t(a) \Bigg[\hat{b}_t(a) + Z_x \lambda_t^{x\top}\left(\frac{B}{2T} \mathbbm{1} - \hat{u}^{x}_{t}(a)\right)\Bigg]\\
\end{align*}
Since $x_{t} \sim q_t^x$, therefore $\E_{\cH_{t-1}}[\sum_{a} q_t^{x}(a)\hat{b}(a)] = \tilde{b}(x_t)$ and $\E_{\cH_{t-1}}[\sum_{a} q_t^{x}(a)\hat{u}^x_t(a)] = {u}^*(x_t)$, therefore using Azuma-Hoeffding with probability at least $1 - \cO\left(\frac{1}{T^2}\right)$

\begin{align}
    \sum_{t=1}^{\tau}\sum_{a} q^{x}_t(a)\hat{\ell}^{x}_t(a) &\leq \sum_{t=1}^{\tau} {b}(x_t) +  Z_x\lambda_t^{x\top}\left( \frac{B}{2T} \mathbbm{1} - u^*(x_t)\right) + \cO(Z_x+1)\sqrt{T \log T}\nonumber
    \label{eq:lhs_termII}
\end{align}
Combining with $\eqref{eq:lhs_termI}$ we get
\begin{align}
   (1-\gamma_x)\sum_{t=1}^{\tau} \hat{\ell}^x_t(a) - \sum_{t=1}^{\tau}\sum_{i=1}^{K} q^x_t(i)\hat{\ell}^x_t(i) &\geq \frac{\tau}{T} \widetilde{\optbx} - \sum_{t=1}^{\tau} {b}(x_t) - Z_x\lambda_t^{x\top}\left( \frac{B}{2T} \mathbbm{1} - u^*(x_t)\right) \nonumber\\
   &- \cO(Z_x+1)\sqrt{T \log T} - \gamma_x (Z_x+1)
\end{align}
From the regret guarantee of OCO on $g^x_{t}(\lambda)$ we have that for any $\lambda \in [0,1]^d$, 
\begin{align}
   (1-\gamma_x)\sum_{t=1}^{\tau} \hat{\ell}^x_t(a) - \sum_{t=1}^{\tau}\sum_{i=1}^{K} q^x_t(i)\hat{\ell}^x_t(i) &\geq \frac{\tau}{T} \widetilde{\optbx} - \sum_{t=1}^{\tau} {b}(x_t) - Z_x\lambda^{\top}\left( \frac{B}{2T} \mathbbm{1} - u^*(x_t)\right) \nonumber\\
   &- \cO(Z_x+1)\sqrt{T \log T} - \gamma_x (Z_x+1) + \cO(\sqrt{T})
    \label{eq:lhs_final}
\end{align}
Next if $\tau = T$, choosing $\lambda = 0$ gives
\begin{align*}
    (1-\gamma_x)\sum_{t=1}^{\tau} \hat{\ell}^x_t(a) - \sum_{t=1}^{\tau}\sum_{i=1}^{K} q^x_t(i)\hat{\ell}^x_t(a) \geq \optb_x - \sum_{t=1}^{\tau} {b}(x_t) - \cO(Z_x+1)\sqrt{T \log T} - \gamma_x (Z_x+1).
\end{align*}
If $\tau < T$ then $\exists j$ such that $\displaystyle \sum_{t=1}^{T} u^*(x_t)_j > B/2$ (i.e., one of the resources is exhausted). Choose $\lambda = Z_x e_j$ and observe that 
\begin{align*}
    \sum_{t=1}^{T} Z_x\lambda^{\top} \left( \frac{B}{2T} \mathbbm{1} - u^*(x_t)\right) &\leq Z_x\left(\frac{\tau}{2T} B - B/2\right)
\end{align*}
Combining with $\eqref{eq:lhs_final}$ we get
\begin{align*}
    (1-\gamma_x)\sum_{t=1}^{\tau} \hat{\ell}^x_t(a) - \sum_{t=1}^{\tau}\sum_{i=1}^{K} q^x_t(i)\hat{\ell}^x_t(i) &\geq \frac{\tau}{T} \widetilde{\optbx} - \sum_{t=1}^{\tau} {b}(x_t) - \frac{\widetilde{2\optb_x}}{B}\left(\frac{\tau}{2T} B - B/2\right) \\
   & -\cO(Z_x+1)\sqrt{T \log T} - \gamma_x (Z_x+1) + \cO(\sqrt{T})\\
    & \geq \widetilde{\optbx} - \sum_{t=1}^{\tau} {b}(x_t) -\cO(Z_x+1)\sqrt{T \log T} - \gamma_x (Z_x+1)
\end{align*}
 \end{proof}
Next we upper bound the RHS of \eqref{eq:exp3gurantee_app} using the following lemma.
 
\lemmatermtwo*
\begin{proof}
Consider the following term:
\begin{align*}
    \sum_{t=1}^{\tau} \sum_{a=1}^{K} {q}^x_{t}(a) \hat{\ell}^x_{t}(a)^2 &=  \sum_{t=1}^{\tau} \sum_{a=1}^{K} {q}^x_{t}(a) \left( \hat{s}_t(a) + Z_x\lambda_t^{x\top}\left( \frac{B}{2T}\mathbbm{1} - \hat{u}_{t}(x_t)\right) \right)^2\\
    &\leq \sum_{t=1}^{\tau} \sum_{a=1}^{K} {q}^x_{t}(a) \left(\hat{s}_t(a)^2 + \left[Z_x\lambda_t^{x\top}\left( \frac{B}{2T}\mathbbm{1} - \hat{u}_{t}(x_t)\right)\right]^2\right)\\
    &\leq \sum_{t=1}^{\tau} \sum_{a=1}^{K} {q}^x_{t}(a) \left(\hat{s}_t(a)^2 + \left[Z_x\lambda_t^{x\top}\left( \frac{B}{2T}\mathbbm{1} - \hat{u}_{t}(x_t)\right)\right]^2\right)\\
    &\leq \sum_{t=1}^{\tau} \sum_{a=1}^{K} {q}^x_{t}(a) \hat{s}_t(a)^2 + \sum_{t=1}^{\tau} \sum_{a=1}^{K} 4Z_x^2 \frac{B^2}{4T^2} + Z_x^2 \sum_{t=1}^{\tau} \sum_{a=1}^{K} 4{q}^x_{t}(a)[\lambda_t^{x\top}\hat{u}^x_t(a)]^2
\end{align*}
We have $\sum_{a}\E[{q}^x_{t}(a) \hat{\ell}^x_{t}(a)^2] \leq \frac{K}{\gamma_x}$ and $\sum_{a}\E {q}^x_{t}(a) [\lambda_t^{x\top}\hat{u}_t(x_t)]^2 \leq \frac{K}{\gamma_x}$ (see \cite[Lemma 6]{pmlr-v139-saha21a}, \cite[Chapter 11]{lattimore_szepesvári_2020}). Using Azuma-Hoeffding, with probability at least $1 - \cO(\frac{1}{T^2})$
\begin{align*}
    \sum_{t=1}^{\tau} \sum_{a=1}^{K} {q}^x_{t}(a) \hat{\ell}^x_{t}(a)^2 &\leq \frac{K}{\gamma_x} T + \frac{Z_x^2 B^2}{T^2} \frac{K}{\gamma_x}T  + \frac{K}{\gamma_x}Z_x^2 T \\
    &\leq \frac{K}{\gamma_x}T (Z_x^2 + 2)
\end{align*}
Therefore 
\begin{align*}
    \frac{\log K}{\eta_x} + \eta_x \sum_{t=1}^{\tau}\sum_{i=1}^{K} {q}^x_t(i)(\hat{\ell}^x_t(i))^2 
    \leq \frac{\log K}{\eta_x} + \eta_x \frac{K}{\gamma_x}T (Z_x^2+2)
\end{align*}
Choosing $\gamma_x = \sqrt{\eta_x K}$ we get
\begin{align*}
    \frac{\log K}{\eta_x} + \eta_x \sum_{t=1}^{\tau}\sum_{i=1}^{K} {q}^x_t(i)(\hat{\ell}^x_t(i))^2 
    \leq \frac{\log K}{\eta_x} + T\sqrt{\eta_x K}(Z_x^2 + 2)
\end{align*}
Finally choosing $\eta_x = \big(\frac{\log K}{T\sqrt{K}}\big)^{2/3}\frac{1}{2Z_x + 1}$ we have
\begin{align*}
    \frac{\log K}{\eta_x} + \eta_x \sum_{t=1}^{\tau}\sum_{i=1}^{K} {q}^x_t(i)(\hat{\ell}^x_t(i))^2 
    &\leq T^{2/3} (K\log K)^{1/3} (2Z_x + 1) + T(Z_x^2+1) \sqrt{\Bigg(\frac{\log K}{T\sqrt{K}}\Bigg)^{2/3}\frac{1}{2Z_x + 1} K}\\
    &\leq \cO\left((K \log K)^{1/3}T^{2/3}Z_x)\right) = \cO\left(\Big(\frac{\optb_x}{B}+1\Big)(K \log K)^{1/3}T^{2/3}\right)
\end{align*}
\end{proof}
\item[] \textbf{Step-3:} We repeat the same argument for the second arm choice $y_t$ and then combining with Lemma~\ref{lemm:lemmaoptshifted} and Lemma~\ref{lemm:lemmaoptmarginal} we get
\begin{align*}
    \widetilde{\optbx} + \widetilde{\optby} - \left(\sum_{t=1}^{\tau} {b}(x_t) + {b}(y_t)\right) \leq  \cO\left(\Big(\frac{\optb_x}{B}+1\Big)(K \log K)^{1/3}T^{2/3}\right)
\end{align*}
which completes the proof.
\end{enumerate}
\end{proof}

\section{Details of Experiments}
\label{sec:exp_app}
We provide more detailed descriptions of our experiments in this section.

\textbf{Datasets.} We run our experiments on two datasets.

\begin{enumerate}
    \item[\textbf{1.}] \textbf{Synthetic Data:} We create a Constrained Dueling Bandits instance with $K = 6$ arms where the preference matrix is given by
    \begin{align*}
        P = \begin{pmatrix}
        0.5& 0.55& 0.55& 0.54& 0.61& 0.61\\
        0.45& 0.5&  0.55& 0.55& 0.58& 0.6 \\
        0.45& 0.45& 0.5&  0.54& 0.51& 0.56\\
        0.46& 0.45& 0.46& 0.5&  0.54& 0.5\\
        0.39& 0.42& 0.49& 0.46& 0.5&  0.51\\
        0.39& 0.4&  0.44& 0.5&  0.49& 0.5 
    \end{pmatrix}.
    \end{align*}
The vector of Borda scores $\Bar{b} = \begin{pmatrix} b(1)& b(2)6& \ldots& b(6)\end{pmatrix}^{\top}$ is given by
$$
\begin{pmatrix}
0.672& 0.646& 0.602& 0.582& 0.554& 0.544
\end{pmatrix}^{\top}.
$$
We experiment with three choices of consumptions. In all three cases the number of resources $d=1$ and the true consumptions across both arms choices are given by the same function, i.e., $u^* = v^*$, and we add zero mean gaussian noise to each entry. The vector of consumptions for arms $\bar{u}^* =\begin{pmatrix} u^*(1)& u^*(2)6& \ldots& u^*(6)
\end{pmatrix}^{\top}$ are given by:
\begin{enumerate}
    \item $\begin{pmatrix}
    0.9&0.9&0.1&0.8&0.8&0.8
    \end{pmatrix}^{\top}$
    \item $\begin{pmatrix}
    0.6&0.5&0.4&0.3&0.2&0.1
    \end{pmatrix}^{\top}$
    \item $\begin{pmatrix}
    0&0&0&0&0&0
    \end{pmatrix}^{\top}$
\end{enumerate}

In the first case although arm 1 and 2 have high Borda scores, the associated consumptions are also high. In the second case the order of consumptions is the same as the order of Borda scores. In the last case all the consumptions are zero and our objective is to evaluate if our algorithm under performs in the absence of constraints.
The experiments are run for $T = 2000$ rounds with $B=1000$ and are run independently over 50 samples.

    \item[\textbf{2.}] \textbf{Car preference dataset:} We consider the Car preference dataset from \cite{Abbasnejad2013} where the preference matrix is generated by considering the user preferences for various models of cars. The dataset utilized 10 items to generate all 45 possible preferences. The study was conducted in two phases, with data collected from 40 and 20 users separately. Participants in the initial experiment were presented with cars featuring specific attributes given by (1) Body type, (2) Transmission, (3) Engine capacity and (4) Fuel consumed. We use the dataset to compute the preference matrix. As in case 1, we consider three choices of consumptions that follow a similar structure as given below:
    \begin{enumerate}
        \item$ \begin{pmatrix}
            0.9&0.9&0.01&0.02&0.7&0.3&0.6&0.7&0.7&0.8
        \end{pmatrix}$
        \item $\begin{pmatrix}
            0.7&0.9&0.9&0.8&0.6&0.1&0.4&0.3&0.5&0.2
        \end{pmatrix}$
        \item $\begin{pmatrix}
            0&0&0&0&0&0&0&0&0&0
        \end{pmatrix}$
    \end{enumerate}
The experiments are run for $T = 5000$ rounds with $B=4000$ and are run independently over 50 samples.
\end{enumerate}

\end{document}